\documentclass{article}
\usepackage{float}
\usepackage{siunitx}

\usepackage[preprint]{neurips_2026}
\usepackage{algorithm}
\usepackage{algpseudocode}
\usepackage[utf8]{inputenc}
\usepackage[T1]{fontenc}

\usepackage[table]{xcolor}
\definecolor{linkblue}{HTML}{005EB8}
\definecolor{OursRowBg}{HTML}{EBF2FA}
\usepackage[colorlinks=true,
  linkcolor=linkblue,
  citecolor=linkblue,
  urlcolor=linkblue]{hyperref}
\usepackage{url}

\usepackage{booktabs}
\usepackage{amsfonts}
\usepackage{amsmath}
\usepackage{amssymb}
\usepackage{nicefrac}
\usepackage{microtype}
\usepackage{xcolor}
\usepackage{graphicx}
\usepackage{multirow} 
\usepackage{cleveref}
\usepackage[most]{tcolorbox}
\usepackage{tikz}
\usetikzlibrary{positioning, arrows.meta, decorations.pathreplacing, fit, shapes.geometric, calc, backgrounds}
\usepackage{pgfplots}
\pgfplotsset{compat=1.18}
\usepackage{etoc}
\tcbset{
  ourqbox/.style={
    enhanced jigsaw,
    breakable,
    colback=black!6!white,
    colframe=black!35!white,
    boxrule=0.5pt,
    arc=2.5pt,
    left=10pt, right=10pt, top=8pt, bottom=8pt,
  }
}

\definecolor{blockbg}{HTML}{DCE9F6}
\definecolor{layerbg}{HTML}{FAE0DC}
\definecolor{blockdark}{HTML}{1F6FB2}
\definecolor{layerdark}{HTML}{C0392B}

\definecolor{blockbg}{HTML}{D8E5F3}    
\definecolor{layerbg}{HTML}{F4DAD8}    
\definecolor{blockdark}{HTML}{1F4F8A}  
\definecolor{layerdark}{HTML}{8B2A2A}  
 
\newcommand{\HiB}[1]{\colorbox{blockbg}{\strut #1}}
\newcommand{\HiL}[1]{\colorbox{layerbg}{\strut #1}}
 
\algrenewcommand\algorithmicrequire{\textbf{Input:}}
\algrenewcommand\algorithmicensure{\textbf{Output:}}
\algrenewcommand\algorithmiccomment[1]{\hfill\textit{\(\triangleright\)\,#1}}

\crefname{algorithm}{Algorithm}{Algorithms}
\Crefname{algorithm}{Algorithm}{Algorithms}

\newcommand{\defeq}{:=}
\newcommand{\eqdef}{=:}
\newcommand{\norm}[1]{\left\lVert#1\right\rVert}

\DeclareMathOperator*{\argmin}{arg\,min}

\newcommand{\ind}{\mathbb{I}}

\newcommand{\Par}[1]{\left(#1\right)}

\title{Training-Free Looped Transformers}

\author{%
  Lizhang Chen\thanks{Equal contribution} \And
  Jonathan Li\footnotemark[1] \And
  Chen Liang \And
  Ni Lao \And
  Qiang Liu
}

\begin{document}
\maketitle

\begingroup
  \renewcommand{\thefootnote}{}
  \footnotetext{University of Texas at Austin}
  \addtocounter{footnote}{-1}
\endgroup

\begin{abstract}
We introduce \textbf{training-free looped transformers}, in which a lightweight inference-time wrapper loops a contiguous mid-stack block of layers of a frozen checkpoint without additional fine-tuning, continued training, or architectural changes. Unlike prior looped transformer methods that train with the looped structure end-to-end, we retrofit recurrence onto pretrained models at test time. We show that naive block reapplication usually degrades performance, highlighting the importance of the loop application strategy. Motivated by \emph{viewing a pre-norm transformer block as a forward Euler step on an ODE}, we instead treat looping as a refinement of the same approximation, replacing one large update with smaller damped sub-steps. Across seven dense, sparse MoE, and MLA+MoE model families, our method improves Qwen3-4B-Instruct by +2.64 pp on MMLU-Pro, Qwen3-30B-A3B-Instruct by +1.14 pp on CommonsenseQA, and Moonlight-16B-A3B-Instruct by +1.20 pp on OpenBookQA.
\end{abstract}

\section{Introduction}
\label{sec:intro}

\emph{Looped transformers} \citep{liu2024looped, yang2025looped}, Universal Transformers \citep{dehghani2019universal,lan2020albert}, and Deep Equilibrium models \citep{bai2019deep,bai2020multiscale} all incorporate \emph{recurrence} into the architecture \citep{xu2025expressive,saunshi2025reasoning,merrill2025little,gong2025whatmakes} and weights \citep{geiping2025scaling,zhu2025ouro,prairie2026parcae,wu2025parallel}. Layers are tied across loop iterations and optimized accordingly, so the recurrence structure is inseparable from the trained parameters. As a consequence, looping cannot be applied to a model that was not trained with this methodology in mind, which is the case for almost all publicly released language-model checkpoints. This naturally gives rise to the core question of this paper:

\smallskip
\begin{tcolorbox}[ourqbox]
\itshape
Can we loop a frozen, off-the-shelf checkpoint directly at inference time, with \textbf{no fine-tuning}, \textbf{no continued training}, \textbf{no auxiliary parameters}, and \textbf{no architectural changes}?
\end{tcolorbox}
\smallskip

Concretely, modern open-weight LLMs, such as Qwen3 \citep{qwen3}, Llama-3.2 \citep{llama3}, Moonlight \citep{moonlight}, and DeepSeek-V2-Lite \citep{deepseekv2}, are the endpoints of a multi-stage pipeline that typically includes continued pretraining, supervised fine-tuning, and one or more rounds of RLHF / DPO post-training \citep{deepseek2025r1}. We take whatever weights the model authors release and apply the loop wrapper at inference, with no further weight updates of any kind.

Several converging strands of evidence point towards a positive answer to core question. \cite{men2024short} show that whole mid-layer blocks of released transformer LMs can be deleted with minimal loss; \cite{lad2024remarkable} report that mid-layers can be skipped, swapped, or repeated without catastrophic degradation; and \cite{belrose2023eliciting} demonstrate that intermediate-layer logits already encode much of the final prediction \citep{takase2021lessons,reid2021subformer}. The prevailing interpretation of these results is one of \emph{compressibility}---middle layers are redundant, so we can remove them---but this ignores a complementary perspective: the same redundancy that makes a mid-layer safe to delete makes it safe to \emph{re-apply}. This insight admits a numerical analysis interpretation that we develop in Section~\ref{sec:strategy}: each pre-norm transformer layer is exactly one forward Euler step at $h{=}1$ on a per-block residual ODE, so \emph{re-applying the block at inference is, geometrically, a finer integration of the ODE the network already approximates} \citep{bai2019deep,bai2020multiscale}. In particular, $K$ sub-steps of size $h{=}1/K$ better approximate the same $t{=}1$ endpoint that the unmodified network was trained to deliver.

We show that this geometric motivation translates into measurable gains on modern checkpoints across \emph{seven model families}---Qwen3 \citep{qwen3} (0.6B/1.7B/4B base \& instruct, 30B-A3B MoE), Qwen1.5-MoE-A2.7B, Llama-3.2 \citep{llama3} (1B/3B), Moonlight \citep{moonlight}, and DeepSeek-V2-Lite \citep{deepseekv2}---and \emph{45 (model, benchmark) cells} of training-free loop evaluation. Looped evaluation yields its strongest gains on knowledge-heavy multiple-choice benchmarks: +2.64 on MMLU-Pro \citep{wang2024mmlupro} and +2.01 on GPQA-Main \citep{rein2024gpqa} for Qwen3-4B-Instruct, and +2.30 on ARC-Challenge \citep{clark2018arc} for Qwen1.5-MoE-A2.7B-Chat. Approximately 20,000 NVIDIA H100 HBM3-80GB GPU hours were used for the experiments.

These improvements arise without any parameter updates, additional supervision, or benchmark-specific tuning, suggesting that repeated application of existing transformer blocks can expose latent inference-time computation. We further compare forward Euler integration with higher-order fixed-point accelerators and solvers, including Anderson acceleration, heavy-ball, Aitken acceleration \citep{walker2011anderson}, and Runge--Kutta-style updates.

Our main contributions are as follows:
\begin{enumerate}
\item \textbf{A training-free loop wrapper}, fully specified, with block and layer iteration modes and seven loop-iteration strategies, comprising of well-known numerical integration methods for an ODE that transformers implicitly approximate.
\item \textbf{Cross-architecture validation} on 7 model families and 45 (model, benchmark) cells under a single \emph{out-of-the-box recipe} ($K$-stage Runge–Kutta at the mid 4 layers; block-mode for dense and layer-mode for MoE) with \emph{no per-cell hyperparameter tuning of any kind}.
\item \textbf{Layer-mode iteration for MoE}: $L_b^K \circ \cdots \circ L_a^K$ rather than $(L_b \circ \cdots \circ L_a)^K$ is necessary for Mixture-of-Experts checkpoints \citep{csordas2024moeut,bae2024relaxed,bae2025mor}, where block-mode causes expert routing to thrash between iterations.
\end{enumerate}

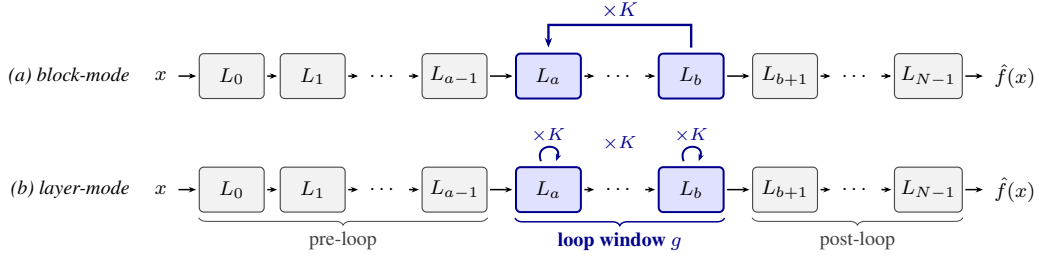
\begin{figure}[t]
\centering
\resizebox{\textwidth}{!}{
\begin{tikzpicture}[
  layer/.style={draw, rounded corners=1.8pt, minimum height=18pt,
                minimum width=26pt, font=\footnotesize, inner sep=2.5pt},
  preloop/.style={layer, fill=gray!10, draw=gray!55!black, line width=0.45pt},
  loopwin/.style={layer, fill=blue!12, draw=blue!55!black, line width=0.8pt},
  postloop/.style={layer, fill=gray!10, draw=gray!55!black, line width=0.45pt},
  arrow/.style={-{Stealth[length=3pt]}, line width=0.6pt,
                shorten >=1pt, shorten <=1pt},
  loopback/.style={-{Stealth[length=4pt]}, line width=0.95pt,
                   draw=blue!55!black},
  perlayerloop/.style={-{Stealth[length=3pt]}, line width=0.8pt,
                       draw=blue!55!black},
  font=\footnotesize,
  node distance=4pt,
]

\begin{scope}[shift={(0,1.6)}]

\node[preloop] (L0a)   at (0.0, 0)  {$L_0$};
\node[preloop, right=6pt of L0a] (L1a)  {$L_1$};
\node[right=6pt of L1a]      (dotsAa){$\cdots$};
\node[preloop, right=6pt of dotsAa] (Lam1a) {$L_{a{-}1}$};

\node[loopwin, right=11pt of Lam1a]  (Laa)   {$L_a$};
\node[right=6pt of Laa]              (dotsBa){$\cdots$};
\node[loopwin, right=6pt of dotsBa]  (Lba)   {$L_b$};

\node[postloop, right=11pt of Lba]   (Lbp1a) {$L_{b{+}1}$};
\node[right=6pt of Lbp1a]            (dotsCa){$\cdots$};
\node[postloop, right=6pt of dotsCa] (LNa)   {$L_{N{-}1}$};

\node[font=\footnotesize\itshape, anchor=east]
   at ($(L0a.west) + (-0.85, 0)$) {(a) block-mode};

\draw[arrow] ($(L0a.west) + (-0.32,0)$) node[left]{$x$} -- (L0a);
\draw[arrow] (L0a) -- (L1a); \draw[arrow] (L1a) -- (dotsAa);
\draw[arrow] (dotsAa) -- (Lam1a); \draw[arrow] (Lam1a) -- (Laa);
\draw[arrow] (Laa) -- (dotsBa); \draw[arrow] (dotsBa) -- (Lba);
\draw[arrow] (Lba) -- (Lbp1a); \draw[arrow] (Lbp1a) -- (dotsCa);
\draw[arrow] (dotsCa) -- (LNa);
\draw[arrow] (LNa) -- ($(LNa.east) + (0.32,0)$) node[right]{$\hat f(x)$};

\draw[loopback]
   ([yshift=2pt]Lba.north) -- ++(0,0.32) -| ([yshift=2pt]Laa.north)
   node[pos=0.25, above=-1pt, font=\footnotesize\bfseries,
        color=blue!55!black] {$\times K$};
\end{scope}

\begin{scope}[shift={(0,0)}]

\node[preloop] (L0b)   at (0.0, 0)  {$L_0$};
\node[preloop, right=6pt of L0b] (L1b)  {$L_1$};
\node[right=6pt of L1b]      (dotsAb){$\cdots$};
\node[preloop, right=6pt of dotsAb] (Lam1b) {$L_{a{-}1}$};

\node[loopwin, right=11pt of Lam1b]  (Lab)   {$L_a$};
\node[right=6pt of Lab]              (dotsBb){$\cdots$};
\node[loopwin, right=6pt of dotsBb]  (Lbb)   {$L_b$};

\node[postloop, right=11pt of Lbb]   (Lbp1b) {$L_{b{+}1}$};
\node[right=6pt of Lbp1b]            (dotsCb){$\cdots$};
\node[postloop, right=6pt of dotsCb] (LNb)   {$L_{N{-}1}$};

\node[font=\footnotesize\itshape, anchor=east]
   at ($(L0b.west) + (-0.85, 0)$) {(b) layer-mode};

\draw[arrow] ($(L0b.west) + (-0.32,0)$) node[left]{$x$} -- (L0b);
\draw[arrow] (L0b) -- (L1b); \draw[arrow] (L1b) -- (dotsAb);
\draw[arrow] (dotsAb) -- (Lam1b); \draw[arrow] (Lam1b) -- (Lab);
\draw[arrow] (Lab) -- (dotsBb); \draw[arrow] (dotsBb) -- (Lbb);
\draw[arrow] (Lbb) -- (Lbp1b); \draw[arrow] (Lbp1b) -- (dotsCb);
\draw[arrow] (dotsCb) -- (LNb);
\draw[arrow] (LNb) -- ($(LNb.east) + (0.32,0)$) node[right]{$\hat f(x)$};

\draw[perlayerloop]
   ([xshift=-3pt, yshift=2pt]Lab.north)
   .. controls +(-0.10, 0.26) and +(0.10, 0.26) ..
   ([xshift=3pt, yshift=2pt]Lab.north);
\node[font=\scriptsize\bfseries, color=blue!55!black, anchor=south]
   at ($(Lab.north) + (0,0.21)$) {$\times K$};

\draw[perlayerloop]
   ([xshift=-3pt, yshift=2pt]Lbb.north)
   .. controls +(-0.10, 0.26) and +(0.10, 0.26) ..
   ([xshift=3pt, yshift=2pt]Lbb.north);
\node[font=\scriptsize\bfseries, color=blue!55!black, anchor=south]
   at ($(Lbb.north) + (0,0.21)$) {$\times K$};

\node[font=\scriptsize\bfseries, color=blue!55!black, anchor=south]
   at ($(dotsBb.north) + (0,0.24)$) {$\times K$};

\draw[decorate, decoration={brace,amplitude=3pt,mirror,raise=2pt},
      semithick, color=blue!55!black]
   (Lab.south west) -- (Lbb.south east)
   node[midway, below=4pt, font=\footnotesize\bfseries,
        color=blue!55!black]
   {loop window $g$};

\draw[decorate, decoration={brace,amplitude=3pt,mirror,raise=2pt},
      thin, color=gray!55!black]
   (L0b.south west) -- (Lam1b.south east)
   node[midway, below=4pt, font=\footnotesize, color=gray!50!black]
   {pre-loop};

\draw[decorate, decoration={brace,amplitude=3pt,mirror,raise=2pt},
      thin, color=gray!55!black]
   (Lbp1b.south west) -- (LNb.south east)
   node[midway, below=4pt, font=\footnotesize, color=gray!50!black]
   {post-loop};

\end{scope}
\end{tikzpicture}
}
\vspace{-2pt}
\caption{\textbf{Training-free looped transformer wrapper, two iteration modes.} A frozen checkpoint is augmented at inference by re-applying a contiguous mid-block $g = L_b\circ\cdots\circ L_a$ for $K$ iterations before resuming the post-loop layers. No weights are changed and no new parameters are introduced; the cost is $(b-a+1)(K-1)$ extra forward passes through the loop window. \textbf{(a) Block-mode} iterates the whole window $K$ times, $(L_b\circ\cdots\circ L_a)^K$. \textbf{(b) Layer-mode} iterates each window layer $K$ times before passing on, $L_b^K\circ\cdots\circ L_a^K$, which is the safer default on Mixture-of-Experts backbones because it pins per-layer expert routing across iterations (Section~\ref{sec:layer-mode}). Section~\ref{sec:strategy} interprets $g^{(K)}$ as sub-stepping the residual ODE the network already integrates with single-step forward Euler.}
\label{fig:method}
\end{figure}

\section{Training-free looped transformers}
\label{sec:method}

\subsection{Loop wrapper}
\label{sec:wrapper}


\begin{algorithm}[t]
\caption{\HiB{\textbf{\textcolor{blockdark}{Block-mode}}} Runge--Kutta forward pass.}
\label{alg:rk-block}
\begin{algorithmic}[1]
\Require $x$, block range $[a,b]$, number of stages $s$, Runge--Kutta coefficients $\{a_{ij}\}_{1\leq j<i\leq s},\{b_i\}_{i=1}^s$
\State $x \gets \textsc{PreLoop}(x)$ \Comment{embedding, pre-block, etc.}
\State \HiB{$y_0 \gets x$} \Comment{\HiB{initial state for the block}}
\For{$i = 1,\dots,s$} \Comment{iterate over the $s$ Runge--Kutta stages}
  \State $y_i \gets y_0 + \sum_{j<i} a_{ij}k_j$ \Comment{add weighted increments}
  \State \HiB{$k_i \gets (L_b\circ\cdots\circ L_a)(y_i) - y_i$} \Comment{\HiB{residual of the full composed block}}
\EndFor
\State {$x \gets y_0 + \sum_{i=1}^{s} b_ik_i$} \Comment{$s^\text{th}$ order Runge--Kutta update}
\State \Return $\textsc{PostLoop}(x)$ \Comment{layers after the loop block, output head}
\end{algorithmic}

\end{algorithm}
\begin{algorithm}[t]
\caption{\HiL{\textbf{\textcolor{layerdark}{Layer-mode}}} Runge--Kutta forward pass.}
\label{alg:rk-layer}
\begin{algorithmic}[1]
\Require $x$, block range $[a,b]$, number of stages $s$, Runge--Kutta coefficients $\{a_{ij}\}_{1\leq j<i\leq s},\{b_i\}_{i=1}^s$
\State $x \gets \textsc{PreLoop}(x)$ \Comment{embedding, pre-block, etc.}
\State \HiL{\textbf{for} $\ell = a, \ldots, b$ \textbf{do}} \Comment{\HiL{outer loop over each layer}}
  \State\quad \HiL{$y_0 \gets x$} \Comment{\HiL{initial state for every layer}}
  \State \quad \textbf{for} $i = 1, \ldots, s$ \textbf{do} \Comment{iterate over the $s$ Runge--Kutta stages}
    \State \quad \quad $y_i \gets y_0 + \sum_{j<i} a_{ij}k_j$ \Comment{add weighted increments}
    \State \quad \quad\HiL{$k_i \gets L_\ell(y_i) - y_i$} \Comment{\HiL{residual of a single layer $L_\ell$}}
\State \quad \textbf{end for}
  \State \quad {$x \gets y_0 + \sum_{i=1}^{s} b_ik_i$} \Comment{$s^\text{th}$ order Runge--Kutta update}
\State \HiL{\textbf{end for}}
\State \Return $\textsc{PostLoop}(x)$ \Comment{layers after the loop block, output head}
\end{algorithmic}
\end{algorithm}

Let $f = L_{N-1} \circ \cdots \circ L_0$ denote a pretrained transformer with $N$ decoder layers \citep{vaswani2017attention,qwen3,llama3,deepseekv2,moonlight}, where $L_i$ maps a residual stream of shape $\mathbb{R}^{T \times d}$ to itself. Choose a contiguous \emph{loop window} indexed by $[a, b]$ with $0 \le a \le b \le N-1$ and a loop count $K \ge 1$. The window induces an operator \begin{equation}
g := L_b \circ L_{b-1} \circ \cdots \circ L_a,
\qquad g : \mathbb{R}^{T \times d} \to \mathbb{R}^{T \times d}.
\label{eq:block-op}
\end{equation}
The looped wrapper splits the network into pre-loop layers $0, \ldots, a-1$, the looped middle, and post-loop layers $b+1, \ldots, N-1$:
\begin{equation}
\hat f(x) =
\underbrace{\bigl(L_{N-1} \circ \cdots \circ L_{b+1}\bigr)}_{\text{post-loop}}
\;\circ\; g^{(K)} \;\circ\;
\underbrace{\bigl(L_{a-1} \circ \cdots \circ L_0\bigr)}_{\text{pre-loop}}(x),
\label{eq:wrapper}
\end{equation}
where $g^{(K)} : \mathbb{R}^{T \times d} \to \mathbb{R}^{T \times d}$ is a $K$-step iteration of $g$. When $a = 0$ or $b = N-1$, the corresponding boundary composition is the identity. The map $g^{(K)}$ depends on an \emph{iteration mode} (Section~\ref{sec:layer-mode}) and a \emph{loop strategy} (Section~\ref{sec:strategy}).

\subsection{Block-mode versus layer-mode iteration}
\label{sec:layer-mode}

Given the window operator $g = L_b \circ \cdots \circ L_a$ in \eqref{eq:block-op}, there are two natural realizations of $g^{(K)}$, \begin{align}
\text{block-mode:}\quad
   &g^{(K)}_{\operatorname{blk}}(x) = (L_b \circ \cdots \circ L_a)^{K}(x), \label{eq:block-mode}\\
\text{layer-mode:}\quad
   &g^{(K)}_{\operatorname{lyr}}(x) = L_b^{K} \circ L_{b-1}^{K} \circ \cdots \circ L_a^{K}(x).
     \label{eq:layer-mode}
\end{align}
In other words, block-mode iterates the entire window as one unit, whereas layer-mode iterates each layer before passing to the next (Figure~\ref{fig:method}). For dense models \citep{bae2024relaxed,bae2025mor}, the two yield broadly similar quality. For mixture-of-experts (MoE) layers \citep{deepseekv2,moonlight,csordas2024moeut}, however, block-mode becomes unstable, since at iteration $i$ the gating network of every MoE layer in the window sees a slightly perturbed hidden state and routes to a different subset of experts than at iteration $i-1$, and the accumulated routing-induced noise eventually overpowers the intended refinement. On the other hand, layer-mode iteration computes the gating decision once and applies the same expert mixture $K$ times \citep{csordas2024moeut}, avoiding this failure mode and making it the correct default for MoE backbones.

\subsection{Loop strategies}
\label{sec:strategy}

We first motivate several loop strategies with a structural observation. A standard pre-norm transformer layer $L$ \citep{vaswani2017attention} implements
\begin{equation}
L(x) = x + \operatorname{Attn}(\operatorname{LN}_1(x))+\operatorname{MLP}(\operatorname{LN}_2(x + \operatorname{Attn}(\operatorname{LN}_1(x)))).
\label{eq:block}
\end{equation}

The right-hand side is the layer's input plus a residual update. Generalizing this directly to the loop window operator $g$, which may be a single layer $g = L_i$ (the layer-mode case) or a contiguous composition $g = L_b \circ \cdots \circ L_a$ (the block-mode case), we define the \emph{window residual field}
\begin{equation}
F_g(x) := g(x) - x.
\label{eq:F}
\end{equation}

By construction, $g(x) = x + F_g(x)$, which is exactly a forward Euler step with step size $h{=}1$
\begin{equation}
    x_1 = x_0 + h \cdot F_g(x_0)
\end{equation}
on the autonomous ODE \citep{bai2019deep,bai2020multiscale}
\begin{equation}
\dot{x}=F_g(x). 
\label{eq:ode}
\end{equation}
This holds regardless of whether the window contains one layer or many, and the two modes differ only in what $F_g$ unfolds to. In layer-mode, $F_g$ is just the single-layer residual field \[ F_g(x) = \operatorname{Attn}_i(\operatorname{LN}^i_1(x)) + \operatorname{MLP}_i(\operatorname{LN}^i_2(\cdot)). \]
In block-mode, telescoping the unrolled chain (proof in Appendix~\ref{app:block-unroll}) gives
\begin{equation}
F_g(x) = \sum_{i=a}^{b} F_{L_i}(y_i(x)),
\qquad y_a(x) := x,\quad y_{i+1}(x) := y_i(x) + F_{L_i}(y_i(x)),
\label{eq:block-effective-field}
\end{equation}
where $F_{L_i}(z) := L_i(z) - z$ is the layer residual field of layer $i$. In both modes, the post-loop layers $L_{b+1}, \ldots, L_{N-1}$ are trained to receive the approximation of $x(t{=}1)$ on $F_g$ implicitly produced by one application of $g$ and not the trajectory at any other time.

\begin{figure}[t]
\centering
\includegraphics[width=\linewidth]{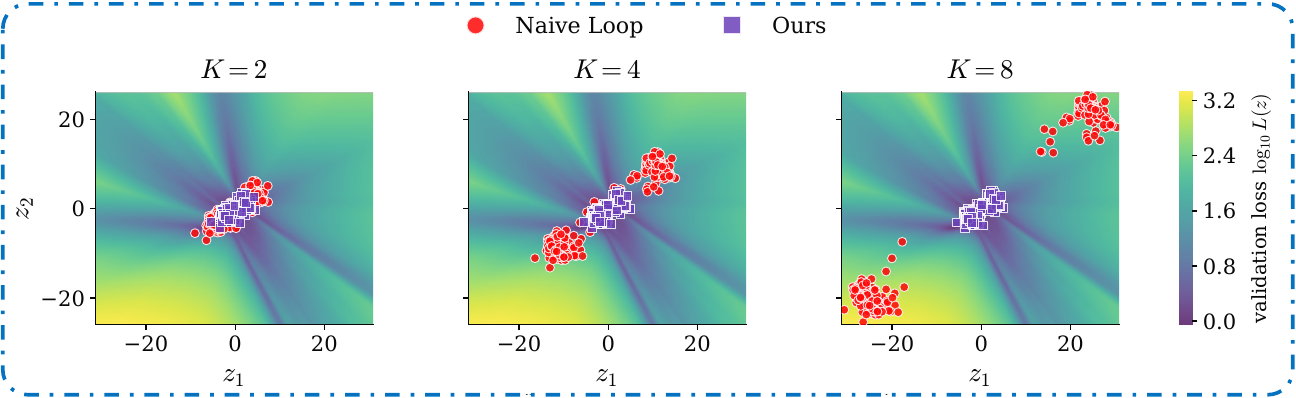}
\caption{\textbf{RK integration vs.\ naive looping.} A tiny MLP $\operatorname{pre}(\mathbb{R}^4\to\mathbb{R}^2)\to$ block of 3 residual layers $(\mathbb{R}^2\to\mathbb{R}^2) \to \operatorname{post}(\mathbb{R}^2\to\mathbb{R}^2)$ is trained end-to-end on a 2-D regression target. Each panel fixes $K \in \{2,4,8\}$ and shows, over a $220{\times}220$ grid in the post-block hidden state $\mathbf{z}$, the median test loss $L(\mathbf{z}) = \operatorname{med}_i \norm{\operatorname{post}(\mathbf{z}) - y_i}^2$ (log scale, colored) together with the test-set scatters obtained by naive looping (red circles) and $K$-stage RK integration (purple squares). The trained baseline induces a central low-loss valley (blue); our endpoints stay clustered in that valley at every $K$, while naive $K$-loop endpoints drift increasingly outward into high-loss regions (yellow) as $K$ increases.}
\label{fig:toy-mlp-contour}
\end{figure}

Naively looping $g$ for $K$ rounds, i.e. $x \leftarrow g(x)$ applied $K$ times \citep{dehghani2019universal,lan2020albert,liu2024looped,yang2025looped,saunshi2025reasoning,xu2025expressive,geiping2025scaling,zhu2025ouro,vonoswald2023transformers,ahn2024transformers,mahankali2024one,gatmiry2024looped,chen2024bypassing}, is a $K$-step forward Euler integration of the window residual field with step size $h{=}1$, which approximates $x(t{=}K)$. But the post-loop layers are not trained to receive the trajectory at $t{=}K$, and empirically, naive $K$-fold looping degrades performance almost universally (Section~\ref{sec:experiments}). Figure~\ref{fig:toy-mlp-contour} illustrates the effect of naive looping vs.\ sub-stepping on a tiny end-to-end trainable network with a $2$-D bottleneck, where the input to the post-loop layers can be directly plotted. The damped sub-step endpoints stay clustered in the trained low-loss valley, while the naive $K$-loop endpoints drift into high-loss regions. Numerically, when averaged over the test set, naive $K{=}2$ looping increases the MSE to 2.88 versus the baseline and sub-step values of 0.015 and 0.36, respectively ($\approx 8\times$ gap); at $K{=}8$ this degrades to a MSE of 335 versus the sub-step value of 1.04 ($\approx 320\times$ gap).

The principled goal of $g^{(K)}$ is therefore not to advance integration to $t{=}K$, but to \emph{better approximate the same endpoint} $x(t{=}1)$ that the unmodified network already targets. To address this challenge, classical numerical analysis suggests to sub-step the same total integration time $[0,1]$ at finer resolution $h = 1/K$. Performing $K$ Euler steps of size $h{=}1/K$ on \eqref{eq:ode} gives the damped update
\begin{equation}
x_{k+1}=\left(1-\frac{1}{K}\right)x_k+\frac{1}{K}g(x_k),
\label{eq:sub-step-euler}
\end{equation}
which converges to the true $x(t{=}1)$ at order $O(1/K)$ and strictly improves upon the single-shot $h{=}1$ Euler step that the network implicitly implements at every layer. Because equations~\eqref{eq:F}--\eqref{eq:ode} are mode-agnostic, this principle applies in both layer-mode (sub-stepping the per-layer field $F_g = F_{L_i}$) and in block-mode (sub-stepping the composite field $F_g$ in~\eqref{eq:block-effective-field}).

More generally, we can consider numerical integration strategies beyond the forward Euler method to approximate $x(t{=}1)$. In particular, the damped Euler update~\eqref{eq:sub-step-euler} can be replaced by the $s$-stage explicit Runge--Kutta integrator
\begin{equation}
x_1 = x_0 + h\sum_{i=1}^{s} b_i k_i,
\label{eq:rk-update}
\end{equation}
where the $s$ stages are given by
\begin{equation}
k_1 = F_g(x_0),\qquad k_2 = F_g(x_0 + h (a_{21} k_1)),\qquad\dots,\qquad k_s = F_g\Par{x_0 + h\sum_{j=1}^{s-1} a_{sj}k_j},
\label{eq:rk-stages}
\end{equation}
with the coefficients $\{a_{ij}, b_i\}$ specified by a Butcher tableau. Algorithms~\ref{alg:rk-block} and~\ref{alg:rk-layer} summarize this family of methods for block-mode and layer-mode implementations, respectively, where the step size is chosen to be $h{=}1$.

\begin{algorithm}[t]
\caption{Explicit \HiB{\textcolor{blockdark}{\textbf{block-mode}}} \HiL{\textcolor{layerdark}{\textbf{layer-mode}}} Runge--Kutta with Butcher tableau \eqref{eq:rk-euler-chain-aij}–\eqref{eq:rk-beta-bi}.}
\label{alg:rk_beta}
\begin{algorithmic}[1]
\Require $x$, range $[a,b]$, number of stages $K$, anchor weight $\beta\in[0,1]$
\State $x \gets \textsc{PreLoop}(x)$ \Comment{embedding, pre-block, etc.}
\State \HiB{$\tilde{x} \gets (L_b \circ \cdots \circ L_a)(x)$} \Comment{\HiB{anchor (block mode)}}
\State \HiL{\textbf{for} $\ell = a, \ldots, b$ \textbf{do}} \Comment{\HiL{outer loop over each layer}}
\State \quad \HiL{$\tilde{x} \gets L_i(x)$} \Comment{\HiL{anchor (layer mode)}}
\State \quad \textbf{for} $k = 1, \ldots, K-1$ \textbf{do} \Comment{iterate over the $K$ Runge--Kutta stages}
\State \quad\quad \HiB{$y \gets (L_b \circ \cdots \circ L_a)(x)$} \HiL{$y \gets L_\ell(x)$}
\State \quad\quad $x \gets x + \frac{1}{K}(y - x)$ \Comment{Runge--Kutta update}
\State \quad \textbf{end for}
\State \quad $x \gets \beta\tilde{x} + (1-\beta)x$ \Comment{anchor combination}
\State \HiL{\textbf{end for}}
\State \Return $\textsc{PostLoop}(x)$
\end{algorithmic}
\end{algorithm}

As a particularly effective example, we choose a specific Butcher tableau so that the RK output becomes an interpolation between the base output and the $K$-step damped Euler updates following~\eqref{eq:sub-step-euler}. Concretely, we set
\begin{equation}
h=1,\quad s=K,\quad\text{and}\quad a_{ij}=\frac{1}{K}\ind(j<i),
\label{eq:rk-euler-chain-aij}
\end{equation}
so that the $i^\text{th}$ stage evaluates $F_g$ at the point
\begin{equation}\label{eq:yi}
y_i=x_0+\frac{1}{K}\sum_{j=1}^{i-1}k_j
    = F^{i-1}(x_0),\qquad k_i=F_g(y_i),
\end{equation}
where $F(x)\defeq x+\frac{1}{K}F_g(x)$ is the damped Euler update~\eqref{eq:sub-step-euler}. We then choose $\beta\in[0,1]$ and set the output weights as
\begin{equation}
b_1=\beta+\frac{1-\beta}{K},
\qquad
b_i=\frac{1-\beta}{K},\quad i=2,\ldots,K,
\label{eq:rk-beta-bi}
\end{equation}
which are nonnegative and sum to one. Under these coefficients, the RK output satisfies the identity
\begin{equation}
x_1=x_0+\sum_{i=1}^{K} b_i k_i=\beta g(x_0)+(1-\beta)F^{K}(x_0),
\label{eq:rk-beta-equivalence}
\end{equation}
which leads to an efficient implementation (Algorithm~\ref{alg:rk_beta}). The proof is provided in Appendix~\ref{app:block-unroll}.

Algorithm~\ref{alg:rk_beta} can be interpreted as a RK method with a front-loaded quadrature rule controlled by $\beta$. In the extreme cases, $\beta=0$ recovers the $K$-substep forward Euler endpoint, while $\beta=1$ preserves the original output $g(x_0)$. For intermediate values of $\beta$, the method effectively biases the trajectory toward the trained one-step endpoint by placing greater weight on the initial residual direction.

\begin{table}[t]
\centering
\caption{\textbf{Iteration strategies for numerical integration of the ODE \eqref{eq:ode}}.}
\small
\setlength{\tabcolsep}{8pt}
\renewcommand{\arraystretch}{1.5}
\resizebox{\textwidth}{!}{%
\begin{tabular}{@{}l l c@{}}
\toprule
\textbf{Strategy} & \textbf{Update Rule} & \textbf{Forward Passes} \\
\midrule
\multicolumn{3}{@{}l}{\textit{Naive iteration \& forward Euler}}\\
\cmidrule(lr){1-3}
Naive Loop
  & $x_{k+1} = g(x_k)$
  & $K$ \\
Forward Euler~\citep{bai2019deep,bai2020multiscale}
  & $x_{k+1} = x_k + \frac{1}{K}F_g(x_k)$
  & $K$ \\
\midrule
\multicolumn{3}{@{}l}{\textit{Higher-order Runge--Kutta}~\citep{butcher2008numerical}}\\
\cmidrule(lr){1-3}
Midpoint (RK2)
  & $x_{k+1} = x_k + hF_g(x_k + \frac{h}{2}F_g(x_k))$
  & $2K$ \\
Heun (RK2)
  & $x_{k+1} = x_k + \frac{h}{2}(F_g(x_k) + F_g(x_k + hF_g(x_k)))$
  & $2K$ \\
RK4
  & $x_{k+1} = x_k + \frac{h}{6}(\kappa_1 + 2\kappa_2 + 2\kappa_3 + \kappa_4)$
  & $4K$ \\
\midrule
\multicolumn{3}{@{}l}{\textit{Fixed-point accelerators}}\\
\cmidrule(lr){1-3}
Heavy-ball ($\alpha,\beta$)~\citep{polyak1964heavyball}
  & $x_{k+1} = x_k + \alpha F_g(x_k) + \beta(x_k - x_{k-1})$
  & $K$ \\
Anderson ($m,\beta$)~\citep{walker2011anderson}
  & $x_{k+1} = (1-\beta)(x_k - (\Delta X_k)\gamma_k^{\star}) + \beta(g(x_k) - (\Delta F_k)\gamma_k^{\star})$
  & $K$ \\
Aitken $\Delta^2$~\citep{aitken1926}
  & $x_{k+1} = x_k - \frac{(d_{1,k})^2}{d_{2,k}}$ \;\;(per-coordinate)
  & $K$ \\
Uniform Loop~\citep{lys2026inner}
  & $x_{k+1} = g\!\left(\frac{1}{k+1}\sum_{i=0}^{k} x_i\right)$
  & $K$ \\
\bottomrule
\end{tabular}}
\scriptsize\raggedright
\emph{Notation.} RK4: $\kappa_1 := F_g(x_k)$,
$\kappa_2 := F_g(x_k + \frac{h}{2}\kappa_1)$,
$\kappa_3 := F_g(x_k + \frac{h}{2}\kappa_2)$,
$\kappa_4 := F_g(x_k + h\kappa_3)$.\\
Anderson:
$\gamma_k^{\star} := \argmin_\gamma\norm{f_k -(\Delta F_k)\gamma}_2$ with $f_i := g(x_i) - x_i$ and $\Delta X_k, \Delta F_k$ the matrices of the last $m$ residual / state increments, respectively \citep{walker2011anderson}.\\
Aitken: $d_{1,k} := g(x_k) - x_k$, $d_{2,k} := g(g(x_k)) - 2g(x_k) + x_k$, applied per coordinate; the Steffensen safeguard clips $|x_{k+1} - x_k| \le |d_{1,k}|$ and requires $K$ even.
\label{tab:strategies}
\vspace{-1em}
\end{table}

Beyond Runge--Kutta methods, there exists a wide range of numerical integration schemes that have been extensively studied in the ODE literature \citep{walker2011anderson,bai2019deep}. These methods are often designed to address specific properties of the underlying dynamics, and we list several prominent examples in Table~\ref{tab:strategies}. These algorithms, which differ in order, accelerator, and step size choices, will form the basis for seven strategies that we evaluate.

\section{Experiments}
\label{sec:experiments}

This section evaluates the proposed training-free loop wrapper across dense and MoE checkpoints, base and instruction-tuned variants, and both standard MHA and MLA backbones, using a fixed recipe with no per-cell hyperparameter tuning (\Cref{sec:setup}), unless otherwise stated.

\paragraph{Main findings.}
Our method yields the largest and most reliable gains on knowledge-heavy multiple-choice tasks, especially MMLU-Pro, GPQA-Main, and ARC-Challenge (\Cref{sec:headline}). Across architectures, most cells are positive or neutral, with failures concentrated in small distilled checkpoints on some knowledge-MC tasks (\Cref{sec:matrix}).
\begin{figure}[t]
\centering
\includegraphics[width=\linewidth]{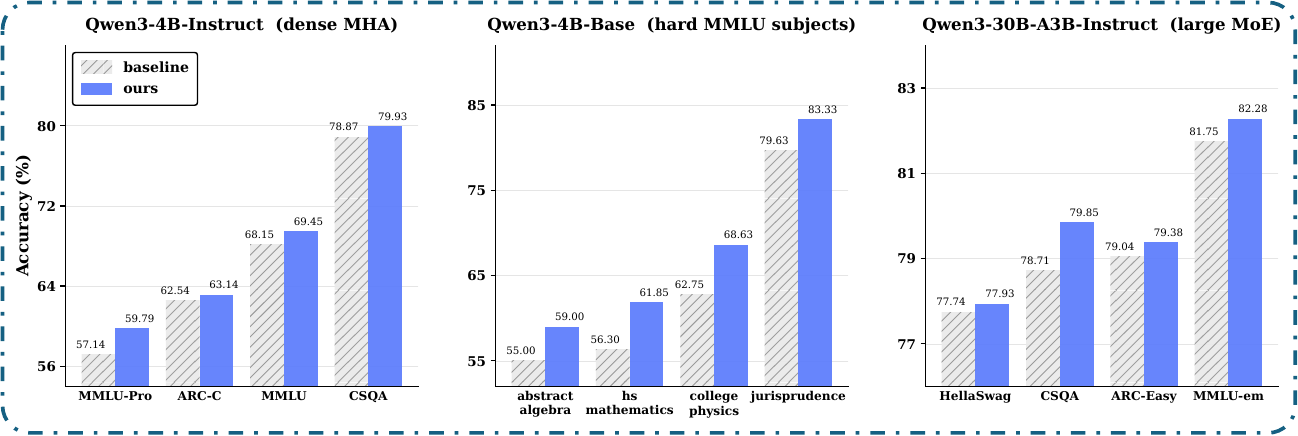}
\caption{\textbf{Per-benchmark accuracy across three Qwen3 model variants under the training-free loop wrapper.} Each panel shows baseline (striped gray) vs.\ our wrapper (solid blue) on 4 knowledge-MC benchmarks. The per-panel y-axis is cropped to emphasize the baseline-to-loop gap. \textbf{Left:} Qwen3-4B-Instruct (dense MHA) on four mid-range general benchmarks. \textbf{Middle:} Qwen3-4B-Base on four hard MMLU 5-shot subjects, selected from MMLU's $57$ subjects per Appendix~\ref{app:per-subject}). \textbf{Right:} Qwen3-30B-A3B-Instruct (large MoE) on four mid-range benchmarks.}
\label{fig:headline-bars}
\end{figure}

\begin{table}[ht]
\centering
\setlength{\aboverulesep}{0pt}
\setlength{\belowrulesep}{0pt}
\caption{\textbf{Window selection for the looping recipe.} \textbf{(a)}~Window-size sweep on Qwen3-1.7B-Base under forward Euler ($K{=}2$): $n{=}4$ is the sweet spot with a sharp cliff at $n{\geq}6$. \textbf{(b)}~The $n{=}4$ choice generalizes across model scales and families. \textbf{(c)}~Block-mode vs.\ layer-mode iteration on MoE backbones at the canonical loop window $[13,16]$ (block $K{=}3$ / layer $K{=}2$): layer-mode yields +0.5 to +1.7 pp on hard MoE benchmarks. Numbers are $\Delta$pp vs.\ baseline. Bold marks the winning variant per row; ``---'' indicates not run.}
\label{tab:window-combined}
 
\begin{minipage}[t]{0.49\linewidth}
\centering
{\small\textbf{(a) Window-size sweep, Qwen3-1.7B-Base}}\par\vspace{3pt}
\scriptsize
\setlength{\tabcolsep}{4pt}
\renewcommand{\arraystretch}{1.18}
\begin{tabular}{@{}c l r r@{}}
\toprule
\textbf{$n$} & \textbf{Window} & \textbf{16-task} & \textbf{$\Delta$} \\
\midrule
1                  & [14]        & 55.72          & +0.18           \\
2                  & [13,14]     & 55.60          & +0.06           \\
3                  & [13,14,15]  & 55.75          & +0.21           \\
\textbf{4} & [12--15]    & \textbf{56.09} & \textbf{+0.55}  \\
6                  & [11--16]    & 54.72          & -0.82           \\
12                 & [8--19]     & 54.91          & -0.63          \\
28 (all)           & [0--27]     & 27.81          & -27.73          \\
\bottomrule
\end{tabular}
\par\vspace{2pt}
 
\vspace{3pt}
 
{\small\textbf{(b) Looped layer sizes across model scales}}\par\vspace{3pt}
\scriptsize
\setlength{\tabcolsep}{4pt}
\renewcommand{\arraystretch}{1.18}
\begin{tabular}{@{}l r r c@{}}
\toprule
\textbf{Model} & \textbf{$n{=}3$ $\Delta$} & \textbf{$n{=}4$ $\Delta$} & \textbf{Winner} \\
\midrule
Qwen3-0.6B-Base~[\citenum{qwen3}]            & \textbf{+0.57} & +0.22           & $n{=}3$ \\
Qwen3-1.7B-Base~[\citenum{qwen3}]            & +0.21          & \textbf{+0.55}  & $n{=}4$ \\
Qwen3-4B-Base~[\citenum{qwen3}]              & +0.43          & \textbf{+0.85}  & $n{=}4$ \\
Llama-3.2-3B-Instruct~[\citenum{llama3}]     & +0.23          & \textbf{+0.45}  & $n{=}4$ \\
\bottomrule
\end{tabular}
\end{minipage}\hfill
\begin{minipage}[t]{0.49\linewidth}
\centering
{\small\textbf{(c) Block-mode vs.\ layer-mode (MoE)}}\par\vspace{3pt}
\scriptsize
\setlength{\tabcolsep}{4pt}
\renewcommand{\arraystretch}{1.18}
\begin{tabular}{@{}l c c@{}}
\toprule
\textbf{Benchmark} & \textbf{block} ($K{=}3$) & \textbf{layer} ($K{=}3$) \\
\midrule
\multicolumn{3}{@{}l}{\textit{Qwen1.5-MoE-A2.7B~[\citenum{qwen15moe}]}}\\
\cmidrule(lr){1-3}
ARC-Challenge & +0.17          & \textbf{+0.85}           \\
CSQA          & +0.16          & \textbf{+0.33}           \\
OBQA          & \textbf{+1.00} & -1.40                    \\
MMLU          & -1.18          & \textbf{\phantom{+}0.00} \\
SciQ          & \textbf{+0.20} & ---                   \\
GPQA-Main     & -1.11          & ---                   \\
\midrule
\multicolumn{3}{@{}l}{\textit{Moonlight-16B-A3B~[\citenum{moonlight}]}}\\
\cmidrule(lr){1-3}
ARC-Challenge & -1.45          & \textbf{+0.51} \\
CSQA          & -0.66          & \textbf{+0.49} \\
OBQA          & -0.20          & \textbf{+1.20} \\
MMLU          & \textbf{+0.74} & ---         \\
SciQ          & \textbf{-0.50} & -0.60          \\
GPQA-Main     & -2.00          & \textbf{+0.90} \\
\bottomrule
\end{tabular}
\end{minipage}
\vspace{-0.1cm}
\end{table}
\paragraph{Robustness.}
The effective loop window follows a stable depth-fraction rule, while MoE models require layer-mode iteration to avoid routing thrash (\Cref{sec:depth}, \Cref{tab:window-combined}). Numerous ablation studies confirm robustness to loop strategy, loop count, window placement, cache handling, and decoding settings (\Cref{sec:ablation-summary}).

\paragraph{Comparisons and transfer.}
Compared with naive inference-time looped transformers, our method avoids collapse and performs best across tested cells (\Cref{sec:method-comparison}, \Cref{fig:method-comparison}). A leakage-free transfer to Qwen3-30B-A3B-Instruct further supports cross-architecture generalization (\Cref{tab:scale-30b}).

\subsection{Experimental setup}
\label{sec:setup}

We evaluate seven Hugging Face checkpoint families spanning dense and MoE backbones, base and instruction-tuned variants, and standard MHA vs.\ Multi-head Latent Attention (MLA): Qwen3 dense $\{0.6,1.7,4\}$B (base \& instruct)~\citep{qwen3}, Qwen3-MoE 30B-A3B-Instruct (\texttt{qwen3\_moe})~\citep{qwen3}, Qwen1.5-MoE-A2.7B-Chat (\texttt{qwen2\_moe}, distilled), Llama-3.2 $\{1,3\}$B-Instruct (distilled from Llama-3.1-8B/70B)~\citep{llama3}, DeepSeek-V2-Lite-Chat (MLA + 64-expert MoE)~\citep{deepseekv2}, and Moonlight-16B-A3B-Instruct (\texttt{deepseek\_v3} family, MLA + MoE)~\citep{moonlight}.

All evaluations use \texttt{lm-eval-harness} 0.4.11 \cite{eval-harness} with \texttt{bfloat16} weights and SDPA attention. For every (model, benchmark) cell, we apply a single \emph{out-of-the-box recipe}: $3$-stage Runge–Kutta at the mid 4 layers, block-mode for dense backbones and layer-mode for MoE backbones, with no per-cell hyperparameter search over position, $K$, or strategy. Per-cell variations in absolute layer indices and in secondary settings (cache, decode) follow mechanically from each architecture's layer count and are logged in Appendix~\ref{app:per-cell} for reproducibility; a fully leakage-free transfer of the recipe \emph{with no per-cell variation of any kind} to a held-out architecture (Qwen3-30B-A3B-Instruct) is reported in Table~\ref{tab:scale-30b}.

\begin{table}[t]
\centering
\caption{\textbf{Knowledge-heavy MC benchmarks with the loop wrapper applied out-of-the-box to each frozen checkpoint.} \emph{No per-cell hyperparameter tuning}. $\Delta$pp is improvement over the no-loop baseline at the same prompt. Per-cell configurations are listed in Appendix~\ref{app:per-cell}, and a fully leakage-free single-recipe generalization check on Qwen3-30B-A3B-Instruct is reported in Table~\ref{tab:scale-30b}.}
\label{tab:headline}
{\small\textbf{(a) Dense backbones}}\label{tab:headline-dense}\par
\small
\setlength{\tabcolsep}{10pt}
\renewcommand{\arraystretch}{1.20}
\setlength{\aboverulesep}{0pt}
\setlength{\belowrulesep}{0pt}
\resizebox{\textwidth}{!}{
\begin{tabular}{l l c >{\columncolor{OursRowBg}}c c}
\toprule
\textbf{Model} & \textbf{Benchmark} & \textbf{Base} & \textbf{Loop (Ours)} & $\boldsymbol{\Delta}$pp \\
\midrule
\multirow{3}{*}{Qwen3-4B-Instruct~[\citenum{qwen3}]}
  & MMLU-Pro 5-shot~[\citenum{wang2024mmlupro}] & 0.5714 & \textbf{0.5979} & \textbf{+2.64} \\
  & GPQA-Main 0-shot~[\citenum{rein2024gpqa}]   & 0.3371 & \textbf{0.3571} & \textbf{+2.01} \\
  & CommonsenseQA 7-shot~[\citenum{talmor2019csqa}] & 0.7887 & 0.7993      & +1.06 \\
\midrule
\multirow{3}{*}{Llama-3.2-3B-Instruct~[\citenum{llama3}]}
  & GPQA-Main 0-shot~[\citenum{rein2024gpqa}]              & 0.2991 & 0.3103 & +1.12 \\
  & MMLU-Pro 5-shot~[\citenum{wang2024mmlupro}]            & 0.3164 & 0.3236 & +0.71 \\
  & MMLU 0-shot~[\citenum{hendrycks2021mmlu}]              & 0.5966 & 0.6039 & +0.72 \\
\midrule
Llama-3.2-1B-Instruct~[\citenum{llama3}]
  & GPQA-Main 0-shot~[\citenum{rein2024gpqa}]              & 0.2790 & 0.2969 & +1.79 \\
\bottomrule
\end{tabular}
}

\vspace{0.6em}
{\small\textbf{(b) MoE backbones}}\label{tab:headline-moe}\par
\small
\setlength{\tabcolsep}{10pt}
\renewcommand{\arraystretch}{1.20}
\resizebox{\textwidth}{!}{
\begin{tabular}{l l c >{\columncolor{OursRowBg}}c c}
\toprule
\textbf{Model} & \textbf{Benchmark} & \textbf{Base} & \textbf{Loop (Ours)} & $\boldsymbol{\Delta}$pp \\
\midrule
\multirow{3}{*}{Qwen1.5-MoE-A2.7B~[\citenum{qwen15moe}]}
  & ARC-Challenge 25-shot~[\citenum{clark2018arc}]    & 0.4829 & \textbf{0.5060} & \textbf{+2.30} \\
  & CommonsenseQA 7-shot~[\citenum{talmor2019csqa}]   & 0.7961 & 0.8133          & +1.72 \\
  & OpenBookQA 0-shot~[\citenum{mihaylov2018obqa}]    & 0.3160 & 0.3320          & +1.60 \\
\midrule
\multirow{2}{*}{Moonlight-16B-A3B~[\citenum{moonlight}]}
  & MMLU 0-shot~[\citenum{hendrycks2021mmlu}]         & 0.6786 & 0.6860 & +0.74 \\
  & OpenBookQA 0-shot~[\citenum{mihaylov2018obqa}]    & 0.3160 & 0.3280 & +1.20 \\
\midrule
DeepSeek-V2-Lite-Chat~[\citenum{deepseekv2}]
  & ARC-Challenge 25-shot~[\citenum{clark2018arc}]    & 0.5794 & 0.5879 & +0.85 \\
\bottomrule
\end{tabular}
}
\end{table}

\subsection{The depth fraction rule}
\label{sec:depth}

Across all eight tested architectures, the optimal window's center sits in a narrow band of fractional depth (Table~\ref{tab:window-combined}, and visualized across nine architectures in Figure~\ref{fig:depth-rule} of Appendix~\ref{app:depth-figure}). For models larger than 1.7B, the optimum lies in the upper half (0.43--0.71, mode $\approx$0.50); for sub-1B models, it shifts earlier (0.25--0.56). We hypothesize that head specialization concentrates in late layers, which the loop must avoid~\citep{belrose2023eliciting,lad2024remarkable,men2024short}; in small or heavily distilled models, the late-layer fraction is larger, so the safe window starts earlier~\citep{takase2021lessons,reid2021subformer}.

\subsection{Layer-mode for MoE}
\label{sec:moe-layer}

On MoE backbones~\citep{deepseekv2,moonlight}, default block-mode iteration causes \emph{routing thrash}, in which each block iteration re-evaluates the gating function on slightly perturbed states, accumulating routing-induced rather than representation-induced changes. This can be fixed by switching to layer-mode iteration, as discussed in Section~\ref{sec:layer-mode} and \cite{csordas2024moeut,bae2024relaxed}. Table~\ref{tab:window-combined} compares the two modes on Qwen1.5-MoE and Moonlight, and layer-mode can be seen to flip most negative cells positive, yielding +0.5 to +1.7 pp on hard MoE benchmarks.

\subsection{Results on knowledge-based MC tasks}
\label{sec:headline}

Our main experimental results concern \emph{knowledge-heavy multiple-choice tasks where the baseline is below ceiling, evaluated with a lenient extractor.} On this class of cells the loop acts as a \emph{frozen-context knowledge refiner}, and the gains are largest and most robust. Figure~\ref{fig:headline-bars} and Table~\ref{tab:headline} report the best loop configuration per model family, separated into dense and MoE backbones.

The largest gains (+2.0 to +2.6 pp) appear on the hardest tasks (MMLU-Pro~\citep{wang2024mmlupro}, GPQA-Main~\citep{rein2024gpqa}, ARC-Challenge~\citep{clark2018arc}) for the strongest MHA backbones. MLA-based MoE models (DeepSeek-V2-Lite, Moonlight) move in the same direction but with 3--4$\times$ smaller magnitudes.

\begin{figure}[t]
    \centering
    \includegraphics[width=\linewidth]{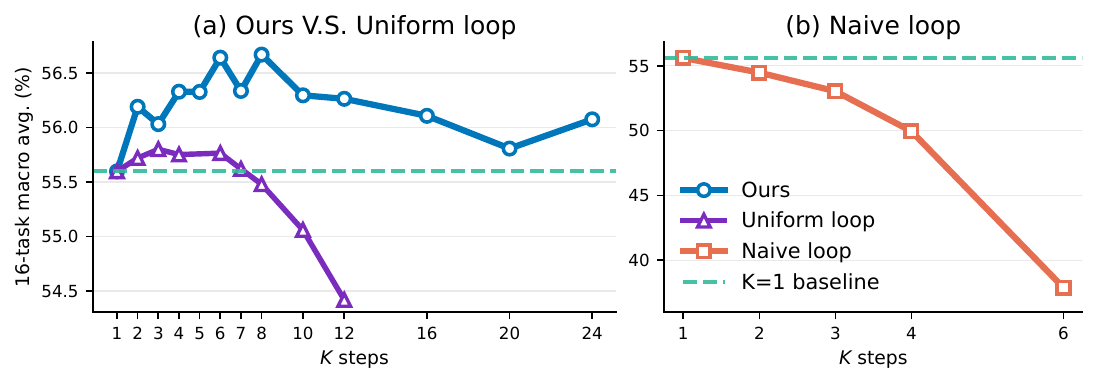}
    \caption{\textbf{Effect of loop count $K$ on 16-task macro-average accuracy.} \textbf{(a)} Ours (Algorithm~\ref{alg:decode-loop-layer-detailed}) is stable across $K \in \{1,2,\cdots, 24\}$, while uniform loop~\cite{lys2026inner} (Table~\ref{tab:strategies}) fails to scale after $K{\geq}6$. \textbf{(b)} The naive loop variant degrades monotonically, falling to 37.89\% at $K{=}6$ (-17.71 pp).}
    \label{fig:k_ablation}
\end{figure}
\begin{table}[t]
\centering
\caption{\textbf{Qwen3-30B-A3B-Instruct generalization check at $[22,24]$.} The configuration is fixed in advance from the depth-fraction rule of Section~\ref{sec:depth}, with \emph{no per-cell search or tuning}, i.e.\ the configuration was committed before any cell was scored, so each cell is a fully leakage-free transfer (Appendix~\ref{app:scale30b}). Bold marks the better of two committed solver choices ($K{=}2$ stage Runge–Kutta in Algorithm \ref{alg:rk_beta} vs.\ Heun $K{=}1$, defined in Table~\ref{tab:solver-ablation}).
}
\label{tab:scale-30b}
\setlength{\tabcolsep}{4.2pt}
\renewcommand{\arraystretch}{1.12}
\resizebox{\textwidth}{!}{
\begin{tabular}{@{}lrrrrrrrrr@{}}
\toprule
& \multicolumn{8}{c}{Accuracy / exact-match ($\uparrow$)}
& \multicolumn{1}{c}{PPL ($\downarrow$)} \\
\cmidrule(lr){2-9}\cmidrule(l){10-10}
Method
& ARC-Easy
& HellaSwag
& SciQ
& CSQA
& TruthfulQA
& MMLU-flex
& GPQA-Main
& SuperGPQA
& LAMBADA \\
\midrule
Baseline
& 79.04 & 77.74 & 94.80 & 78.71 & 34.15 & 66.67 & 37.28 & 31.00 & 4.12 \\
\midrule
\rowcolor{OursRowBg}
 $K$-stage RK
& \textbf{79.38} & \textbf{77.93} & \textbf{95.00} & \textbf{79.85}
& \textbf{34.64} & \textbf{67.46} & 37.50 & \textbf{31.70} & \textbf{4.11} \\
$\Delta$ vs. baseline
& +0.34 & +0.19 & +0.20 & +1.14
& +0.49 & +0.79 & +0.22 & +0.70 & -0.01 \\
\midrule
\rowcolor{OursRowBg}
Heun $K{=}1$
& 79.08 & 77.78 & 94.90 & 79.61
& 34.27 & 66.67 & \textbf{37.72} & 31.05 & 4.12 \\
$\Delta$ vs. baseline
& +0.04 & +0.04 & +0.10 & +0.90
& +0.12 & 0.00 & +0.44 & +0.05 & 0.00 \\
\bottomrule
\end{tabular}}
\vspace{-0.5em}
\end{table}

\subsection{Cross-architecture summary}
\label{sec:matrix}

Across the seven model families and a uniform knowledge-MC task suite, we score every cell under the fixed recipe of Section~\ref{sec:setup}, resulting in 27 positive ($\Delta>+0.3$ pp, 60\%) and 12 neutral ($|\Delta|\le0.3$ pp, 27\%) cells. The remaining negatives concentrate in a single regime, namely \emph{sub-3B distilled checkpoints on knowledge MC} (e.g. Llama-3.2-1B at multiple-choice MMLU~\citep{hendrycks2021mmlu}). Even in this regime, GPQA-Main~\citep{rein2024gpqa} \emph{does} flip positive on Llama-3.2-1B (+1.79) at very-early position $[4,7]$, showing the failure boundary is task-dependent rather than absolute. The wrapper produces a positive or neutral signal on \emph{87\% of cells} across dense, MoE, base,
instruct, and distilled checkpoints.

\subsection{Ablations}
\label{sec:ablation-summary}

We systematically conducted numerous ablation studies, with full tables deferred to the appendix. These include the integration strategy and damping schedule (Appendix~\ref{app:strategy-ablation}), loop count $K$ (Figure~\ref{fig:k_ablation}), window width and position (Appendices~\ref{app:failures} and~\ref{app:position-landscape}), block-mode versus layer-mode iteration (Appendix~\ref{app:layer-mode-dense}), KV-cache and decode-time handling (Appendices~\ref{app:cache-robustness} and~\ref{app:wallclock}), and recipe-robustness checks varying window position, strategy, iteration count, and cache choice (Appendices~\ref{app:per-cell} and~\ref{app:search-protocol}). We further report robustness checks, per-subject decompositions, large-scale untuned transfer, and failed-configuration analyses in Appendices~\ref{app:robustness}, \ref{app:per-subject}, \ref{app:scale30b}, and~\ref{app:failures}.

\subsection{Comparison with other looped transformer methods}
\label{sec:method-comparison}

\begin{figure}[t]
\centering
\includegraphics[width=\linewidth]{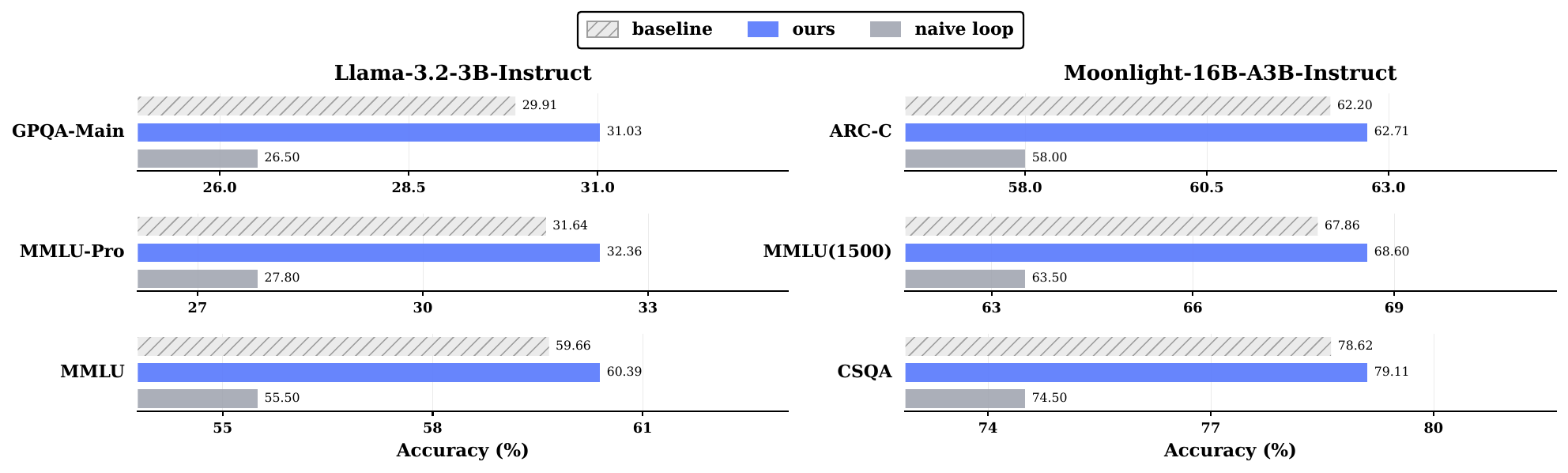}
\caption{\textbf{Comparison with other looped transformer methods on Llama-3.2-3B-Instruct and Moonlight-16B-A3B-Instruct.} On each backbone we report three knowledge-MC benchmarks under three configurations: \emph{baseline} (no loop, original checkpoint), \emph{naive loop} with $K{=}4$, and \emph{ours} (Algorithm~\ref{alg:rk_beta}).}
\vspace{-1em}
\label{fig:method-comparison}
\end{figure}

We further isolate the contribution of our method by comparing it against two natural alternatives on the same frozen checkpoints~\citep{dehghani2019universal,yang2025looped,geiping2025scaling,zhu2025ouro,bae2024relaxed,bae2025mor,jeddi2026loopformer,oncescu2026recurrent,tang2026looprpt,saunshi2025reasoning,xu2025expressive,wu2025parallel}. The naive looped transformer of \cite{liu2024looped} collapses on every cell, since the iterates leave the regime the post-loop layers were trained on (Section~\ref{sec:strategy}), and accuracy drops by several
percentage points. In comparison, our method remains within the trained regime, producing the highest accuracy on nearly every cell. Figure~\ref{fig:method-comparison} reports the results on Llama-3.2-3B-Instruct (GPQA-Main~\citep{rein2024gpqa}, MMLU-Pro~\citep{wang2024mmlupro}, MMLU~\citep{hendrycks2021mmlu}) and Moonlight-16B-A3B-Instruct (ARC-C~\citep{clark2018arc}, MMLU, CSQA~\citep{talmor2019csqa}).

\section{Related work}
\label{sec:related}

\paragraph{Looped and recurrent-depth transformers.}
A large body of work incorporates recurrence into the transformer at \emph{training} time, from weight-tied stacks \citep{dehghani2019universal,lan2020albert} and looped expressivity results \citep{liu2024looped,xu2025expressive,merrill2025little,gong2025whatmakes, YangLNP24, GeipingYS25} to length-generalization studies \citep{yang2025looped,deluca2024simulation,schwarzschild2021canyou,bansal2022endtoend} and looped pretraining at scale \citep{geiping2025scaling,zhu2025ouro,prairie2026parcae,tang2026looprpt,wu2025parallel}. Adaptive-recursion variants \citep{gao2024algoformer,csordas2024moeut,bae2025mor,bae2024relaxed,nouriborji2025modernalbert} and implicit-depth fixed points \citep{bai2019deep,bai2020multiscale} likewise tie recurrence to training, as do recent elastic-depth latent reasoning models \citep{jeddi2026loopformer,oncescu2026recurrent,chen2026thinking,knupp2026depth,yu2026spiralformer,pappone2025twoscale,koishekenov2025encode}. Our wrapper instead targets \emph{unmodified released checkpoints} \citep{qwen3,llama3,deepseekv2,moonlight} with no training \citep{LiangLCL24, liu2024distributed, liang2024cautious, nguyen2024memory, chen2025demystifying, chen2025cautious, NguyenLCL26}, no auxiliary loss, and no architectural changes.

\paragraph{Inference-time compute and latent reasoning.}
A separate axis spends test-time compute on longer trajectories, e.g.\ chain-of-thought \citep{wei2022chain, MohtashamiPJ25}, self-consistency \citep{wang2023selfconsistency}, and reasoning-trained LMs \citep{deepseek2025r1}, all leaving the per-token forward unchanged; theory characterizes their advantages \citep{feng2023revealing,merrill2024expressive,xu2025tocot,saunshi2025reasoning}. Latent-reasoning lines instead push compute into a continuous space \citep{hao2024coconut,cheng2024compressed,pfau2024letsthink,goyal2024think,zeng2025ponder,yu2025relay,geiping2025scaling, li2026learning} or use adaptive halting to ``ponder'' \citep{graves2016adaptive,banino2021pondernet,kohli2026loop,lu2025latentcot,liang2026latentcot,chen2026loopbridge}, echoing the tuned-lens view of forward computation as iterative refinement \citep{belrose2023eliciting} and layer-redundancy findings \citep{men2024short,lad2024remarkable,takase2021lessons,reid2021subformer}. Our intervention is orthogonal, as we use more compute \emph{per token within a single forward pass} on a frozen checkpoint.

\paragraph{Numerical analysis methods.}
Treating a residual block as a forward Euler step motivates the family of integration strategies our wrapper exposes on the looped window, in line with the fixed-point picture \citep{bai2019deep,bai2020multiscale}. Within that family we benchmark Picard acceleration via Anderson \citep{walker2011anderson}, heavy-ball \citep{polyak1964heavyball, ChenLLL24, ChenLL25, peng2026demo, LiaoYSZCL26, su2026the}, Aitken extrapolation, and higher-order Runge--Kutta solvers, and find none robustly beats $K$-stage Runge–Kutta, suggesting the looped middle of a pretrained transformer \citep{liu2024looped,saunshi2025reasoning,bae2025mor} is not contractive in any useful sense \citep{bai2019deep,belrose2023eliciting}. Concurrent training-free directions manipulate depth via layer skipping \citep{men2024short,lad2024remarkable}, but to our knowledge, no prior method combines mid-block looping and layer-mode iteration for MoE routing \citep{ChenLWLLLLL26} across modern checkpoints \citep{qwen3,llama3,deepseekv2,moonlight}.

\section{Conclusion}
\label{sec:conclusion}

Our training-free wrapper applies to frozen released checkpoints (Qwen3, Llama-3.2, DeepSeek-V2-Lite, Moonlight, Qwen1.5-MoE) by reinterpreting each pre-norm transformer block as a forward Euler step at $h{=}1$ and providing a better approximation via iterating a contiguous loop window $K$ times. Block-mode and layer-mode iteration are both supported, with layer-mode required on MoE backbones to stabilize per-layer routing. A universal depth-fraction rule places the optimal window at fractional depth $\approx$0.45--0.60 across dense and MoE architectures from 16 to 48 layers. Gains concentrate on knowledge-heavy multiple-choice benchmarks (MMLU-Pro \citep{wang2024mmlupro}, GPQA-Main \citep{rein2024gpqa}, ARC-Challenge \citep{clark2018arc}), where our method adds roughly +2 pp, and 87\% of cells across seven model families and our benchmark suite are non-negative under the fixed out-of-the-box recipe with no per-cell hyperparameter tuning.

\bibliographystyle{plain}
\bibliography{references}

\clearpage
\appendix

\etocsetlocaltop.toc{part}
\begingroup
  \etocsetnexttocdepth{subsection}
  \etocsettocstyle{\section*{Appendix}}{}
  \localtableofcontents
\endgroup
\clearpage
\section{Notation and glossary}
\label{app:notation}

Table~\ref{tab:notation} consolidates the symbols used across the paper and the appendix.

\begin{table}[h]
\centering
\footnotesize
\setlength{\tabcolsep}{6pt}
\renewcommand{\arraystretch}{1.10}
\caption{Symbols, types, and where they are introduced.}
\label{tab:notation}
\begin{tabular}{@{}llp{0.55\linewidth}@{}}
\toprule
Symbol & Type / range & Meaning \\
\midrule
\multicolumn{3}{@{}l}{\textit{Network}}\\
$N$                                & $\mathbb{N}_{>0}$
                                   & Number of decoder layers in the released checkpoint. \\
$L_i$                              & $\mathbb{R}^{T\times d}\!\to\!\mathbb{R}^{T\times d}$
                                   & Pre-norm transformer block $i$, $0 \le i \le N-1$. \\
$f$                                & composition
                                   & Unmodified network: $f = L_{N-1} \circ \cdots \circ L_0$. \\
$T,d$                            & $\mathbb{N}_{>0}$
                                   & Sequence length and hidden dimension. \\
\addlinespace
\multicolumn{3}{@{}l}{\textit{Loop wrapper}}\\
$[a,b]$                            & $0 \le a \le b \le N-1$
                                   & Loop-window layer indices (contiguous). \\
$W$                                & $b-a+1$
                                   & Loop window width (in layers). \\
$K$                                & $\mathbb{N}_{\ge 1}$
                                   & Loop iteration count. \\
$g$                                & $\mathbb{R}^{T\times d}\!\to\!\mathbb{R}^{T\times d}$
                                   & Block operator: $g = L_b \circ \cdots \circ L_a$
                                     (Eq.~\ref{eq:block-op}). \\
$g^{(K)}$                          & $\mathbb{R}^{T\times d}\!\to\!\mathbb{R}^{T\times d}$
                                   & $K$-step iteration of $g$ under the chosen strategy
                                     (Section~\ref{sec:strategy}). \\
$\hat f(x)$                        & $\mathbb{R}^{T\times d}$
                                   & Looped network output \eqref{eq:wrapper}. \\
$g^{(K)}_{\mathrm{blk}},g^{(K)}_{\mathrm{lyr}}$
                                   & maps
                                   & Block-mode and layer-mode realizations of $g^{(K)}$
                                     \eqref{eq:block-mode}--\eqref{eq:layer-mode}. \\
\addlinespace
\multicolumn{3}{@{}l}{\textit{ODE view}}\\
$F_g$                              & $\mathbb{R}^{T\times d}\!\to\!\mathbb{R}^{T\times d}$
                                   & Window residual field $F_g(x) := g(x) - x$
                                     \eqref{eq:F}. \\
$F_{L_i}$                          & $\mathbb{R}^{T\times d}\!\to\!\mathbb{R}^{T\times d}$
                                   & Single-layer residual field $L_i(x) - x$. \\
$y_i(x)$                           & $\mathbb{R}^{T\times d}$
                                   & Intra-window residual stream at layer $i$
                                     \eqref{eq:block-effective-field}. \\
$h$                                & $\mathbb{R}_{>0}$
                                   & Forward-Euler step size; $h{=}1$ is the layer-built-in step. \\
$\alpha$                           & $(0,1]$
                                   & Damped-Euler coefficient $\alpha$; $\alpha = h$. \\
$\kappa_1,\ldots,\kappa_4$         & $\mathbb{R}^{T\times d}$
                                   & RK4 stage evaluations (Table~\ref{tab:strategies}). \\
$m,\beta$                        & $\mathbb{N},\,\mathbb{R}$
                                   & Anderson memory depth and mixing parameter. \\
\addlinespace
\multicolumn{3}{@{}l}{\textit{KV cache}}\\
$\mathcal{C}_i$                    & cache slot
                                   & Per-layer KV slot for layer $i$. \\
$\ell_i$                           & $\mathbb{N}$
                                   & Snapshot length of $\mathcal{C}_i$ before a loop iteration. \\
$c$                                & $\{\textsc{first},\textsc{last},\textsc{none}\}$
                                   & Cache strategy \eqref{eq:cache-strategy}. \\
\bottomrule
\end{tabular}
\end{table}

\section{Proofs}
\label{app:block-unroll}

\subsection{Proof of \texorpdfstring{\eqref{eq:block-effective-field}}{(9)}}

We expand block-mode iteration as an explicit chain of per-layer forward Euler steps and verify the closed-form identity stated in Section~\ref{sec:strategy}.

Let $g = L_b \circ L_{b-1} \circ \cdots \circ L_a$ be the loop window operator. Each constituent layer satisfies the single-layer forward Euler identity $L_i(z) = z + F_i(z)$. Fix an input $x$. Set $y_a(x) := x$ and define the \emph{intra-window residual stream} recursively by
\begin{equation}
y_{i+1}(x) \defeq L_i(y_i(x))
        \defeq y_i(x) + F_i\bigl(y_i(x)),
\qquad i = a, a+1, \ldots, b.
\label{eq:app-unroll-step}
\end{equation}
By construction, $g(x) = y_{b+1}(x)$. Summing \eqref{eq:app-unroll-step} from
$i = a$ to $i = b$ gives
\begin{equation*}
\sum_{i=a}^{b} (y_{i+1}(x) - y_i(x))
   = \sum_{i=a}^{b} F_i\bigl(y_i(x)).
\end{equation*}
The LHS telescopes to $y_{b+1}(x) - y_a(x) = g(x) - x$. Therefore
\begin{equation}
g(x) - x = \sum_{i=a}^{b} F_i(y_i(x)) \eqdef F_g(x),
\label{eq:app-block-effective-field}
\end{equation}
which is exactly \eqref{eq:block-effective-field}.

\subsection{Proof of \texorpdfstring{\eqref{eq:rk-beta-equivalence}}{(16)}}

By \eqref{eq:yi}, the standard $K$-substep forward Euler endpoint can be written as \[ F^{(K)}(x_0)=x_0+\frac{1}{K}\sum_{i=1}^{K}k_i. \] Substituting into the RHS of \eqref{eq:rk-beta-equivalence}, we have \begin{align*}
    \beta g(x_0)+(1-\beta)F^{(k)}(x_0)&=\beta g(x_0)+(1-\beta)x_0+\frac{1-\beta}{K}\sum_{i=1}^{K}k_i\\
    &=\beta g(x_0)+(1-\beta)x_0+\frac{1-\beta}{K}F_g(x_0)+\frac{1-\beta}{K}\sum_{i=2}^Kk_i\\
    &=x_0+\Par{\beta+\frac{1-\beta}{K}}k_1+\frac{1-\beta}{K}\sum_{i=2}^Kk_i,
\end{align*}
where the second and third lines use $k_1=F_g(x_0)=g(x_0)-x_0$. The conclusion follows upon expanding the LHS of \eqref{eq:rk-beta-equivalence} using the definitions in \eqref{eq:rk-beta-bi}.

\section{Decode loop algorithm}
\label{app:decode-loop}

The looped forward pass we patch into a frozen model has two regimes that share most code but differ in how the loop body interacts with the KV cache. During \emph{prefill} (\texttt{seq\_len > 1}, no past KV) the loop body is run with \texttt{use\_cache=False} and writes nothing to the cache; only a single \emph{stash pass} after the loop writes the canonical KV. During \emph{decode} (\texttt{seq\_len == 1}, an existing past KV) the loop body must attend to the past KV; otherwise, the new token is computed against a truncated context. However, it must also produce no net KV writes, since the stash pass will write the canonical entry afterwards. The mechanism is to snapshot per-loop-layer cache lengths before each loop iteration and crop back to those lengths immediately after.

\subsection{Decode-time looped forward pass}
\label{app:decode-algo-detailed}

Algorithm~\ref{alg:decode-loop-detailed} states the full block-mode decode-time forward pass as a single procedure, including the pre-loop / loop-body / stash / post-loop phases, the per-iteration snapshot/restore protocol, and the cache-strategy branch. Algorithm~\ref{alg:decode-loop-layer-detailed} states the layer-mode counterpart, which differs only in how iterations and snapshot/restore are nested.

\begin{figure}[t]
\centering
\begin{algorithm}[H]
\caption{Decode-time block-mode looped forward pass with KV cache}
\label{alg:decode-loop-detailed}
\begin{algorithmic}[1]
\Require new-token hidden state $x \in \mathbb{R}^{1 \times d}$, loop window $[a,b]$, loop count $K$, strategy update $\mathcal{S}$ (Euler/heavy-ball/Anderson/$\ldots$), cache strategy $c \in \{\textsc{first},\textsc{last},\textsc{none}\}$, per-layer KV cache $\mathcal{C} = (\mathcal{C}_0,\ldots,\mathcal{C}_{N-1})$
\Ensure Updated hidden state $x'$; $\mathcal{C}$ contains exactly one canonical KV entry per loop layer at the new decode position

\For{$i = 0,\ldots,a-1$} \Comment{pre-loop: standard layers, 1 KV each}
    \State $x \gets L_i(x;\mathcal{C}_i,\mathtt{use\_cache}=\textsc{T})$
\EndFor

\State $x_a \gets x$ \Comment{save pre-loop input for $c=\textsc{first}$}

\For{$i = a,\ldots,b$} \Comment{snapshot loop-layer cache lengths}
    \State $\ell_i \gets |\mathcal{C}_i|$
\EndFor

\For{$k = 1,\ldots,K$} \Comment{$K$ loop iterations under strategy $\mathcal{S}$}
    \State $y \gets x$ \Comment{evaluation starts from $x$}
    \For{$i = a,\ldots,b$} \Comment{run body with $\mathtt{use\_cache}=\textsc{T}$}
        \State $y \gets L_i(y;\mathcal{C}_i,\mathtt{use\_cache}=\textsc{T})$ \Comment{attn reads past KV at decode pos.}
    \EndFor
    \For{$i = a,\ldots,b$} \Comment{crop: discard iteration KV writes}
        \State $\mathcal{C}_i \gets \mathtt{crop}(\mathcal{C}_i,\ell_i)$ \Comment{net cache effect of body $=0$}
    \EndFor
    \State $x \gets \mathcal{S}(x,y,k)$ \Comment{e.g. $x+\alpha(y-x)$ for damped Euler}
\EndFor

\If{$c \neq \textsc{none}$} \Comment{stash phase: 1 KV write per loop layer}
    \State $z \gets \begin{cases}
        x, & c=\textsc{last},\\
        x_a, & c=\textsc{first}
    \end{cases}$
    \For{$i = a,\ldots,b$}
        \State $z \gets L_i(z;\mathcal{C}_i,\mathtt{use\_cache}=\textsc{T})$ \Comment{canonical KV at decode pos.}
    \EndFor
\EndIf

\For{$i = b+1,\ldots,N-1$} \Comment{post-loop: standard layers, 1 KV each}
    \State $x \gets L_i(x;\mathcal{C}_i,\mathtt{use\_cache}=\textsc{T})$
\EndFor

\State \Return $\textsc{LayerNorm}_{\text{out}}(x)$
\end{algorithmic}
\end{algorithm}
\end{figure}

\begin{figure}[t]
\centering
\begin{algorithm}[H]
\caption{Decode-time \emph{layer-mode} looped forward pass with KV cache (per-layer iteration; the safer default on MoE backbones, Section~\ref{sec:layer-mode}).}
\label{alg:decode-loop-layer-detailed}
\begin{algorithmic}[1]
\Require Same as Algorithm~\ref{alg:decode-loop-detailed}.
\Ensure Same as Algorithm~\ref{alg:decode-loop-detailed}.

\For{$i = 0, \ldots, a-1$} \Comment{pre-loop, standard}
    \State $x \gets L_i(x;\,\mathcal{C}_i,\,\mathtt{use\_cache}{=}\textsc{T})$
\EndFor

\State $x_a \gets x$

\For{$i = a, \ldots, b$} \Comment{outer: layers in order}
    \State $\ell_i \gets |\mathcal{C}_i|$ \Comment{snapshot length for layer $i$}
    \For{$k = 1, \ldots, K$} \Comment{inner: iterate this layer $K$ times}
        \State $y \gets L_i(x;\,\mathcal{C}_i,\,\mathtt{use\_cache}{=}\textsc{T})$
        \Comment{MoE: routing pinned across $k$}
        \State $\mathcal{C}_i \gets \mathtt{crop}(\mathcal{C}_i,\,\ell_i)$
        \Comment{discard iteration KV write}
        \State $x \gets \mathcal{S}(x,\,y,\,k)$
        \Comment{per-layer strategy update}
    \EndFor
    \State $z \gets x$ \textbf{if} $c{=}\textsc{last}$ \textbf{else} $x_a$
    \If{$c \neq \textsc{none}$}
        \State $z \gets L_i(z;\,\mathcal{C}_i,\,\mathtt{use\_cache}{=}\textsc{T})$
        \Comment{canonical KV at layer $i$}
    \EndIf
\EndFor

\For{$i = b+1, \ldots, N-1$} \Comment{post-loop, standard}
    \State $x \gets L_i(x;\,\mathcal{C}_i,\,\mathtt{use\_cache}{=}\textsc{T})$
\EndFor

\State \Return $\textsc{LayerNorm}_{\text{out}}(x)$
\end{algorithmic}
\end{algorithm}
\end{figure}

Lines 4--5 of Algorithm~\ref{alg:decode-loop-detailed} record the cache length each loop layer holds \emph{before} the body runs. The body in lines 6--12 is then executed $K$ times: each iteration runs the window with \texttt{use\_cache=True} (so attention reads the genuine past KV at the new decode position, line 9), and the immediately-following crop in lines 10--11 truncates the cache back to $\ell_i$, leaving zero net KV effect from the iteration. The strategy step on line 12 produces the next iterate $x$ from the current input $x$ and the body output $y$. Lines 13--16 are the stash phase: a single additional pass through the loop layers writes exactly one canonical KV entry per loop layer per decode position, using either the post-loop hidden state ($c{=}\textsc{last}$) or the pre-loop input ($c{=}\textsc{first}$) as the input to that pass. Lines 17--19 finish with the standard post-loop layers and the final output norm. The total cache delta of the entire procedure is exactly $b - a + 1$ entries regardless of $K$, which is identical to the unmodified model.

\subsection{Snapshot/restore}

\texttt{transformers.DynamicCache.update()} unconditionally appends the new $(k,v)$ to its per-layer tensor: there is no in-place overwrite mode. A naive $K$-iteration decode body would therefore write $K$ extra KV entries per loop layer at the same logical decode position, then attend to all of them on the next decode step, corrupting the cache and producing a sequence of phantom prefix tokens. The snapshot/restore pattern is the cheapest fix that (a) lets each loop iteration attend to the genuine past KV and (b) leaves the cache exactly as it was pre-iteration. The single canonical KV is then written by the stash pass on the chosen \texttt{cache\_strategy}. \texttt{layer-mode} looping uses an analogous per-layer snapshot/crop pair, applied $K$ times to one layer before moving on.

\subsection{\texttt{decode\_mode} variants}

We expose three \texttt{decode\_mode} settings. \texttt{bypass} skips the loop entirely on incremental decode and only loops during prefill (the default for loglikelihood-only evaluations). \texttt{full} loops every decode step. \texttt{first\_n} loops only on the first $n$ generated tokens (intended for CoT-prefix refinement; in practice never beats \texttt{full} when the loop helps and never recovers when it hurts). \texttt{full} is the default for the generation results reported in the main paper.

\section{Per-cell configurations}
\label{app:per-cell}

Table~\ref{tab:per-cell} lists every (model, benchmark) cell that is non-negative ($\Delta \geq -0.3$~pp) under the fixed recipe of Section~\ref{sec:setup}, i.e. $K{=}2$-stage Runge--Kutta at the depth-fraction window, block-mode for dense backbones and layer-mode for MoE, with \emph{no per-cell hyperparameter tuning}. Per-cell variations in absolute layer indices and in secondary settings (cache, decode) follow mechanically from each architecture's layer count; see Appendix~\ref{app:search-protocol} for the recipe specification. Each row gives the no-loop baseline, the best loop configuration found for that cell, and the resulting $\Delta$ in percentage points. Configurations are written as \texttt{[start--end] mode K strategy cache decode}, e.g.\ \texttt{[15--18] block K=3 Euler first full}.

\begin{table}[h]
\centering
\scriptsize
\setlength{\tabcolsep}{3pt}
\caption{Per-cell loop configurations across 7 model families under the out-of-the-box recipe of Section~\ref{sec:setup}. Mode = block / layer; cache $\in$ \{first, last, none\}; decode $\in$ \{bypass, full\}. The loop region iterates the indicated layer range $K$ times under the indicated strategy.}
\label{tab:per-cell}
\resizebox{\textwidth}{!}{%
\begin{tabular}{lllrrll}
\toprule
Model & Benchmark & Baseline & Loop & $\Delta$pp & Window / $K$ / strategy & cache / mode \\
\midrule
Qwen3-4B-Instruct  & MMLU-Pro 5sh CoT~[\citenum{wang2024mmlupro}]  & 57.14 & 59.79 & +2.64 & [15--18] $K{=}3$ Euler  & first / block / full \\
Qwen3-4B-Instruct  & GPQA-Main 0sh~[\citenum{rein2024gpqa}]     & 33.71 & 35.71 & +2.01 & [15--18] $K{=}3$ Euler  & first / block / full \\
Qwen3-4B-Instruct  & SciQ 0sh~[\citenum{welbl2017sciq}]          & 93.30 & 95.10 & +1.80 & [15--18] $K{=}3$ Euler  & first / block / full \\
Qwen3-4B-Instruct  & MMLU 0sh~[\citenum{hendrycks2021mmlu}]          & 68.15 & 69.45 & +1.30 & [15--18] $K{=}3$ Euler  & first / block / full \\
Qwen3-4B-Instruct  & CommonsenseQA 7sh~[\citenum{talmor2019csqa}] & 78.87 & 79.93 & +1.06 & [15--18] $K{=}3$ Euler  & first / block / full \\
Qwen3-4B-Instruct  & MedMCQA 0sh~[\citenum{pal2022medmcqa}]       & 53.40 & 54.13 & +0.73 & [11--14] $K{=}3$ Euler  & first / block / full \\
Qwen3-4B-Instruct  & ARC-Challenge 25sh~[\citenum{clark2018arc}]& 62.54 & 63.14 & +0.60 & [15--18] $K{=}3$ damped Euler $\alpha{=}0.5$ halt $\tau{=}0.05$ & first / block / full \\
Qwen3-4B-Instruct  & C-Eval 5sh~[\citenum{huang2023ceval}]        & 72.21 & 72.73 & +0.52 & [15--18] $K{=}3$ Euler  & first / block / full \\
Qwen3-4B-Instruct  & OpenBookQA 0sh~[\citenum{mihaylov2018obqa}]    & 40.20 & 40.60 & +0.40 & [11--14] $K{=}3$ Euler  & first / block / full \\
\midrule
Qwen3-4B-Base      & MMLU 5sh          & 73.27 & 73.61 & +0.34 & [15--18] $K{=}3$ Euler  & first / block / -- \\
Qwen3-4B-Base      & 16-task macro     & 63.98 & 65.03 & +1.05 & [15--18] $K{=}3$ Euler  & first / block / -- \\
Qwen3-1.7B-Base    & MMLU 5sh          & 62.68 & 63.01 & +0.34 & [12--15] $K{=}2$ damped Euler    & last  / block / -- \\
Qwen3-1.7B-Base    & 16-task macro     & 55.60 & 56.11 & +0.51 & [12--15] $K{=}2$ damped Euler    & last  / block / -- \\
Qwen3-0.6B-Base    & 16-task macro     & 47.87 & 48.43 & +0.56 & [12--15] $K{=}2$ heavy-ball $\alpha{=}0.5,\beta{=}0.5$ & last / layer / -- \\
\midrule
Llama-3.2-3B-Inst  & MMLU-Pro 5sh CoT  & --    & --    & +0.71 & [12--15] $K{=}2$ Euler  & first / block / full \\
Llama-3.2-3B-Inst  & GPQA-Main 0sh     & --    & --    & +0.67 & [12--15] $K{=}2$ Euler  & first / block / full \\
Llama-3.2-3B-Inst  & MMLU 5sh          & --    & --    & +0.72 & [16--19] $K{=}2$ Euler  & last  / layer / bypass \\
Llama-3.2-3B-Inst  & OpenBookQA 0sh    & --    & --    & +0.40 & [20--23] $K{=}3$ Euler  & last  / block / bypass \\
\midrule
Qwen1.5-MoE-A2.7B  & ARC-Challenge 25sh& 48.29 & 50.59 & +2.30 & [13--16] $K{=}3$ Euler  & first / block / full \\
Qwen1.5-MoE-A2.7B  & CommonsenseQA 7sh & 79.61 & 81.33 & +1.72 & [10--13] $K{=}2$ Euler  & first / block / full \\
Qwen1.5-MoE-A2.7B  & OpenBookQA 0sh    & 31.60 & 33.20 & +1.60 & [14--17] $K{=}3$ Euler  & first / block / full \\
Qwen1.5-MoE-A2.7B  & SciQ 0sh          & 94.80 & 95.10 & +0.30 & [14--17] $K{=}3$ Euler  & first / block / full \\
\midrule
DeepSeek-V2-Lite   & ARC-Challenge 25sh& 57.94 & 58.79 & +0.85 & [13--16] $K{=}2$ Euler  & first / layer / bypass \\
DeepSeek-V2-Lite   & OpenBookQA 0sh    & 35.80 & 36.60 & +0.80 & [13--16] $K{=}2$ Euler  & first / block / bypass \\
DeepSeek-V2-Lite   & MMLU (1500)       & --    & --    & +0.56 & [10--13] $K{=}3$ Euler  & first / block / bypass \\
DeepSeek-V2-Lite   & CommonsenseQA 7sh & 75.43 & 75.92 & +0.49 & [11--14] $K{=}2$ Euler  & first / block / bypass \\
\midrule
Moonlight-16B-A3B  & OpenBookQA 0sh    & --    & --    & +1.20 & [8--11]  $K{=}3$ Euler  & first / block / bypass \\
Moonlight-16B-A3B  & GPQA-Main 0sh     & --    & --    & +0.89 & [15--18] $K{=}2$ Euler  & first / layer / bypass \\
Moonlight-16B-A3B  & ARC-Challenge 25sh& --    & --    & +0.51 & [11--14] $K{=}2$ Euler  & first / layer / bypass \\
Moonlight-16B-A3B  & CommonsenseQA 7sh & --    & --    & +0.49 & [11--14] $K{=}2$ Euler  & first / layer / bypass \\
\midrule
Qwen3-30B-A3B-Inst & CommonsenseQA 7sh & 78.71 & 79.85 & +1.14 & [22--24] $K{=}2$ Euler  & first / block / bypass \\
\bottomrule
\end{tabular}%
}
\end{table}

We ablate strategy, $K$, window width, and cache strategy on Qwen3-1.7B-Base at canonical position [12-15]. Table~\ref{tab:ablations} reports the results.

\begin{table}[h]
\centering
\caption{Ablations of the loop wrapper on Qwen3-1.7B-Base \texttt{[12--15]}. $\Delta$pp on the $16$-task macro vs.\ the no-loop baseline; the within-family reference cell is damped Euler at $K{=}2,\,\alpha{=}0.5$. $\downarrow\downarrow$ marks catastrophic divergence (perplexity blowup or strict-format regression). Bold marks the per-axis best.}
\label{tab:ablations}
\small
\renewcommand{\arraystretch}{1.15}
\setlength{\tabcolsep}{5pt}
\resizebox{0.92\textwidth}{!}{%
\begin{tabular}{@{}l *{6}{c}@{}}
\toprule
\multicolumn{7}{@{}l}{\textbf{(a) Window width $n$} \,(centered on \texttt{[12--15]}, $K{=}2$ damped Euler)}\\
\midrule
$n$        & 1     & 3     & \textbf{4}        & 6     & 12    & 28 (full)             \\
\cmidrule(lr){2-7}
$\Delta$pp & +0.18 & +0.21 & \textbf{+0.55}    & -0.82 & -0.63 & $\downarrow\downarrow$  \\
\midrule
\multicolumn{7}{@{}l}{\textbf{(b) Iteration count $K$ \& strategy} \,($n{=}4$, mid-window)}\\
\midrule
           & \multicolumn{2}{c}{\textit{naive loop}}
           & \multicolumn{2}{c}{\textit{damped Euler}}
           & \multicolumn{2}{c}{\textit{higher-order \& accel.}} \\
\cmidrule(lr){2-3}\cmidrule(lr){4-5}\cmidrule(lr){6-7}
$K$        & $K{=}2$       & $K{=}4$
           & $K{=}2,\,\alpha{=}0.5$ & $K{=}3,\,\alpha{\approx}1/3$
           & \multicolumn{2}{c}{Anderson, RK4, Heun, $\ldots$} \\
\cmidrule(lr){2-3}\cmidrule(lr){4-5}\cmidrule(lr){6-7}
$\Delta$pp & +0.51       & -10.21$_{\text{(0.6B)}}$
           & \textbf{+0.55} & +1.05$_{\text{(4B)}}$
           & \multicolumn{2}{c}{see Table~\ref{tab:solver-ablation}} \\
\midrule
\multicolumn{7}{@{}l}{\textbf{(c) Cache strategy} \,($n{=}4$, $K{=}2$ damped Euler)}\\
\midrule
           & \multicolumn{3}{c}{\textit{16-task aggregate}}
           & \multicolumn{3}{c}{\textit{per-benchmark detail (\texttt{none} regression)}} \\
\cmidrule(lr){2-4}\cmidrule(lr){5-7}
strategy   & \texttt{first}    & \texttt{last}     & \texttt{none}
           & GSM8K~[\citenum{cobbe2021gsm8k}] \texttt{none} & MMLU-Pro~[\citenum{wang2024mmlupro}] \texttt{none} & MBPP~[\citenum{austin2021mbpp}] $\Delta_{\texttt{last-first}}$ \\
\cmidrule(lr){2-4}\cmidrule(lr){5-7}
$\Delta$pp & $\geq$\textbf{0} & $\geq$\textbf{0} & $\downarrow\downarrow$
           & -58.91          & -8.86           & +0.60              \\
\bottomrule
\end{tabular}%
}
\end{table}

Table~\ref{tab:solver-ablation} pulls representative rows from the $40+$ configuration sweep in Appendix~\ref{app:strategy-ablation}, demonstrating that higher-order integrators do not help. The looped block is not a smooth ODE, and Anderson-style fixed-point acceleration~\citep{walker2011anderson,bai2019deep,bai2020multiscale} overshoots because the block is not contractive enough. Within our strategy family, $K{=}2$-stage Runge--Kutta with $n{=}4$ and \texttt{cache=first} is the robust default.

\begin{table}[h]
\centering
\caption{Higher-order ODE solvers and fixed-point accelerators vs.\ damped Euler. Numbers are 16-task macro $\Delta$pp on Qwen3-1.7B-Base \texttt{[12--15]}; each column is referenced to its own damped Euler $K{=}2,\,\alpha{=}0.5$ baseline. ``---'' = not tested in that mode. The full sweep ($40+$ configurations across both modes) is reported in Appendix~\ref{app:strategy-ablation}.}
\label{tab:solver-ablation}
\small
\setlength{\tabcolsep}{8pt}
\renewcommand{\arraystretch}{1.12}
\begin{tabular*}{\textwidth}{@{\extracolsep{\fill}}llcc@{}}
\toprule
Family & Method ($K{=}2$ unless noted) & Block-mode $\Delta$ & Layer-mode $\Delta$ \\
\midrule
\textit{Reference}
   & damped Euler ($\alpha{=}0.5$, $h{=}1/2$)
   & 0.00 & 0.00 \\
\midrule
\multirow{3}{*}{\textit{Higher-order Runge--Kutta}}
   & midpoint (RK2) & -0.85 & -1.96 \\
   & Heun (RK2)     & -0.92 & -1.90 \\
   & RK4            & -0.65 & -2.34 \\
\midrule
\multirow{4}{*}{\textit{Fixed-point acceleration}}
   & heavy-ball ($\alpha{=}0.5,\,\beta{=}0.3$)
       & -0.12 & -0.06 \\
   & Anderson ($K{=}4,\,m{=}2,\,\beta{=}0.5$)
       & -0.96 & ---      \\
   & Aitken $\Delta^2$ (safeguarded)
       & -7.00 & ---      \\
   & Anderson, worst ($K{=}8,\,m{=}3,\,\beta{=}1.0$)
       & \textbf{-18.06} & \textbf{-19.56} \\
\bottomrule
\end{tabular*}
\end{table}

\subsection{Patterns visible in the per-cell table}

Three patterns generalize across families: (i) the winning window sits at depth fraction 0.4--0.7 for all dense and MoE models tested, with sub-1B models sometimes preferring a much earlier band (e.g.\ Llama-3.2-1B GPQA prefers \texttt{[4--7]} on a 16-layer model); (ii) MoE models systematically prefer \texttt{layer-mode} $K{=}2$ over \texttt{block-mode} $K{=}3$ on individual cells, consistent with expert-routing thrash hurting block-mode at $K \geq 3$; and (iii) \texttt{cache=first} dominates \texttt{cache=last} on long-prompt cells, while \texttt{cache=last} is mildly preferred for short structured generation (MBPP, +0.6 pp over \texttt{first}).

\section{Strategy ablation tables}
\label{app:strategy-ablation}

This section reproduces, in full numerical form, the strategy comparisons that motivate a main result of the paper: \emph{every classical fixed-point acceleration we tried fails to outperform $K{=}2$-stage Runge--Kutta (Algorithm~\ref{alg:rk_beta}) with $\beta{=}0.5$ on the looped block.} All entries use Qwen3-1.7B-Base on the canonical loop window \texttt{[12--15]} with block-mode looping. The reference is $K{=}2$ damped Euler at 0.56113 macro on the 16-task aggregate.

\subsection{Anderson, heavy-ball, Aitken, \texorpdfstring{$\alpha$}{alpha}-schedules}

\begin{table}[h]
\centering
\footnotesize
\caption{Fixed-point acceleration sweep (Qwen3-1.7B-Base, \texttt{[12--15]}, block-mode). $\Delta$ is in percentage points vs the $K{=}2$ damped Euler reference 56.113. Sorted best$\to$worst.}
\label{tab:phase6}
\begin{tabular}{lclcc}
\toprule
Strategy & $K$ & Hyperparameters & 16-task & $\Delta$ vs $K{=}2$ damped Euler \\
\midrule
Euler-sched     & 2 & $\alpha{=}[0.6,0.4]$       & 56.085 & -0.028 \\
heavy-ball      & 2 & $\alpha{=}0.5,\beta{=}0.3$ & 55.999 & -0.115 \\
heavy-ball      & 2 & $\alpha{=}0.5,\beta{=}0.5$ & 55.993 & -0.120 \\
heavy-ball      & 2 & $\alpha{=}0.7,\beta{=}0.3$ & 55.946 & -0.167 \\
heavy-ball      & 4 & $\alpha{=}0.3,\beta{=}0.5$ & 55.927 & -0.186 \\
Euler-sched     & 4 & $\alpha{=}[0.7,0.5,0.3,0.1]$ & 55.478 & -0.636 \\
Euler-sched     & 3 & $\alpha{=}[0.7,0.5,0.3]$    & 55.252 & -0.861 \\
Anderson        & 4 & $m{=}2,\beta{=}0.5$        & 55.153 & -0.960 \\
Euler-sched     & 3 & $\alpha{=}[0.3,0.5,0.7]$    & 55.036 & -1.077 \\
Anderson        & 6 & $m{=}3,\beta{=}0.5$        & 54.722 & -1.391 \\
Anderson        & 3 & $m{=}2,\beta{=}1.0$        & 54.373 & -1.740 \\
Anderson        & 6 & $m{=}2,\beta{=}0.5$        & 54.333 & -1.781 \\
heavy-ball      & 4 & $\alpha{=}0.5,\beta{=}0.3$ & 54.044 & -2.069 \\
Euler-sched     & 4 & $\alpha{=}[0.4,0.6,0.6,0.4]$& 54.005 & -2.108 \\
heavy-ball      & 4 & $\alpha{=}0.5,\beta{=}0.5$ & 53.495 & -2.618 \\
Anderson        & 4 & $m{=}3,\beta{=}1.0$        & 52.655 & -3.458 \\
heavy-ball      & 4 & $\alpha{=}0.5,\beta{=}0.7$ & 52.562 & -3.552 \\
Anderson        & 4 & $m{=}2,\beta{=}1.0$        & 52.032 & -4.082 \\
heavy-ball      & 3 & $\alpha{=}1.0,\beta{=}0.5$ & 51.236 & -4.877 \\
Aitken $\Delta^2$ & 2 & safeguarded             & 49.109 & -7.004 \\
Aitken $\Delta^2$ & 4 & safeguarded             & 48.318 & -7.795 \\
Aitken $\Delta^2$ & 6 & safeguarded             & 44.921 & -11.192 \\
Anderson        & 6 & $m{=}2,\beta{=}1.0$        & 44.261 & -11.852 \\
Anderson        & 8 & $m{=}3,\beta{=}1.0$        & 38.058 & -18.056 \\
\bottomrule
\end{tabular}
\end{table}

24 configurations; \emph{none} beat $K{=}2$ damped Euler. The closest is the $[0.6,0.4]$ damped-Euler schedule at -0.03 pp (statistical tie). Anderson extrapolation degrades smoothly with $K$ and catastrophically at $K{=}8$ (-18 pp), confirming the residual sequence $x_{k+1} - x_k$ in the loop is not approaching zero in a way Anderson can exploit.

\subsection{Norm stabilization and polynomial blending}

\begin{table}[h]
\centering
\footnotesize
\caption{Norm-stab and poly-blend sweep, same setup. $\Delta$
in pp vs $K{=}2$ damped Euler reference; baseline (no loop) sits at $55.598$.}
\label{tab:phase7}
\begin{tabular}{lrlrr}
\toprule
Strategy & $K$ & Hyperparameters & 16-task & $\Delta$ vs $K{=}2$ damped Euler \\
\midrule
norm\_stab  & 2 & $\alpha{=}0.5$               & 56.059 & -0.055 \\
poly\_blend & 2 & $w{=}[0.4,0.2,0.4]$          & 55.974 & -0.139 \\
poly\_blend & 2 & $w{=}[0.1,0.4,0.5]$          & 55.884 & -0.229 \\
poly\_blend & 2 & $w{=}[0.2,0.3,0.5]$          & 55.753 & -0.360 \\
poly\_blend & 2 & $w{=}[0.0,0.5,0.5]$          & 55.599 & -0.514 \\
poly\_blend & 2 & $w{=}[0.25,0.25,0.5]$        & 55.371 & -0.742 \\
poly\_blend & 2 & $w{=}[0.5,0.0,0.5]$          & 55.107 & -1.006 \\
norm\_stab  & 2 & $\alpha{=}0.7$               & 55.009 & -1.105 \\
poly\_blend & 3 & $w{=}[0.2,0.2,0.2,0.4]$      & 54.818 & -1.295 \\
poly\_blend & 3 & $w{=}[0.1,0.2,0.3,0.4]$      & 54.681 & -1.433 \\
norm\_stab  & 3 & $\alpha{=}0.5$               & 53.270 & -2.844 \\
norm\_stab  & 2 & $\alpha{=}1.0$               & 52.302 & -3.811 \\
norm\_stab  & 6 & $\alpha{=}0.3$               & 49.560 & -6.554 \\
norm\_stab  & 3 & $\alpha{=}0.7$               & 49.338 & -6.776 \\
norm\_stab  & 4 & $\alpha{=}0.5$               & 49.067 & -7.046 \\
norm\_stab  & 3 & $\alpha{=}1.0$               & 43.101 & -13.013 \\
\bottomrule
\end{tabular}
\end{table}

16 additional configurations; again none beat $K{=}2$ damped Euler. \texttt{norm\_stab} (rescaling each iterate to a fixed L2 norm) is catastrophic at $K \geq 3$, confirming that the iterates are diverging \emph{in direction} too.

\subsection{Layer-mode replicates the result}

In layer-mode the loop body is $L_i^K$ applied per-layer rather than $g^K = (L_b \circ \cdots \circ L_a)^K$ applied to the block. Reference: layer-mode $K{=}2$ damped Euler at 56.347 on the same window.

\begin{table}[h]
\centering
\footnotesize
\caption{Layer-mode strategy sweep on Qwen3-1.7B-Base \texttt{[12--15]}. $\Delta$ vs layer-mode $K{=}2$ damped Euler.}
\label{tab:phase18}
\begin{tabular}{lrr}
\toprule
Recipe & 16-task & $\Delta$ vs layer $K{=}2$ damped Euler \\
\midrule
$K{=}2$ heavy-ball $\alpha{=}0.5,\beta{=}0.3$ & 56.292 & -0.056 \\
$K{=}2$ poly\_blend $w{=}[0.25,0.5,0.25]$    & 56.181 & -0.167 \\
$K{=}3$ damped Euler $\alpha{=}0.3$           & 55.278 & -1.069 \\
$K{=}4$ Euler                                 & 55.240 & -1.107 \\
$K{=}2$ damped Euler $\alpha{=}0.7$           & 55.196 & -1.151 \\
$K{=}6$ Euler                                 & 54.998 & -1.350 \\
$K{=}2$ Heun                                  & 54.445 & -1.902 \\
$K{=}2$ midpoint                              & 54.385 & -1.962 \\
$K{=}1$ RK4                                   & 54.242 & -2.105 \\
$K{=}1$ midpoint                              & 54.178 & -2.169 \\
$K{=}2$ norm\_stab $\alpha{=}0.5$             & 54.098 & -2.249 \\
$K{=}1$ Heun                                  & 54.060 & -2.287 \\
$K{=}2$ RK4                                   & 54.005 & -2.342 \\
$K{=}3$ heavy-ball $\alpha{=}0.5,\beta{=}0.3$ & 51.830 & -4.517 \\
$K{=}3$ Anderson $m{=}2,\beta{=}1.0$          & 44.035 & -12.312 \\
$K{=}4$ Anderson $m{=}2,\beta{=}1.0$          & 36.785 & -19.562 \\
\bottomrule
\end{tabular}
\end{table}

Higher-order ODE solvers (Heun, midpoint, RK4) degrade more dramatically in layer-mode (-1.7 to -2.3 pp) than in block-mode (-0.7 to -1.1 pp): per-layer the map $L_i$ is \emph{not} better conditioned for higher-order integration than the composed $g$. Anderson in layer-mode is even more severe (-19.6 pp at $K{=}4$). The combined sweep covers 40+ acceleration configurations, and \emph{none} robustly beats Runge--Kutta. This is the empirical basis for the claim that \emph{the looped transformer block is not a contractive map}, so fixed-point acceleration is the wrong tool.

\section{Compute and reproducibility}
\label{app:reproducibility}

\subsection{Software stack}

All evaluations use \texttt{lm-evaluation-harness} v0.4.11 with deterministic seeds, default sampling settings per task, and \texttt{transformers} 4.46. Looped forwards are implemented as monkey patches on the model's \texttt{Model} class; patching preserves all original generation/eval code paths and is reversible. For the DeepSeek-V2/V3 family we additionally ship a small \texttt{DynamicCache} compatibility shim that re-exposes \texttt{seen\_tokens}/\texttt{get\_max\_length}/\texttt{get\_usable\_length} to the bundled \texttt{modeling\_deepseek} remote code.

\subsection{Per-model resource footprint}

\begin{table}[h]
\centering
\footnotesize
\caption{Approximate single-GPU memory and per-job wallclock for the representative jobs. Times are for the full 16-task suite at default \texttt{lm-eval-harness} batch sizes; CoT generation jobs (e.g.\ MMLU-Pro 5sh CoT) take 2--5$\times$ longer in \texttt{decode\_mode=full}.}
\begin{tabular}{lrll}
\toprule
Model & VRAM (bf16) & Hardware used & Eval-suite wallclock \\
\midrule
Qwen3-0.6B-Base / -Instruct       &  $\sim$2~GB & A100-40 / H100-80 & $\sim$1~h \\
Qwen3-1.7B-Base                   &  $\sim$4~GB & A100-40 / H100-80 & $\sim$2~h \\
Qwen3-4B-Base / -Instruct         &  $\sim$8~GB & A100-40 / H100-80 & $\sim$3~h \\
Llama-3.2-1B / -3B-Instruct       &  $\sim$3--7~GB & A100-40 / H100-80 & $\sim$2~h \\
Qwen1.5-MoE-A2.7B-Chat            &  $\sim$30~GB & H100-80 & $\sim$4~h \\
DeepSeek-V2-Lite-Chat (16B/2.4B)  &  $\sim$32~GB & H100-80 & $\sim$5~h \\
Moonlight-16B-A3B-Instruct        &  $\sim$32~GB & H100-80 & $\sim$8~h (\texttt{decode=full} 21~h) \\
Qwen3-30B-A3B-Instruct            &  $\sim$60~GB & H100-80 & $\sim$10~h (19-task 0-shot) \\
\bottomrule
\end{tabular}
\end{table}

\section{Failed configurations log}
\label{app:failures}

This section documents the configurations we tried that broke. We include them in the appendix so future work knows what does \emph{not} work, and so the main claims are read against an honest log of what \emph{was} tried.

\subsection{Catastrophic collapse on naive \texorpdfstring{$K{=}4$}{K=4} looping}

The first ablation we ran (Qwen3-0.6B-Instruct, 6-task) tested naive $K{=}4$ looping ($x \leftarrow g(x)$ four times, no damping). This resulted in lambada perplexity blowing up from $\sim$13 to \textbf{1054.12}, and 16-task macro dropped -10.21 pp. This is the original empirical observation that the block is not a contraction: undamped iteration diverges visibly within four applications, even on a window of just four mid layers.

\subsection{\texttt{cache=none} is uniformly catastrophic on decode}

Setting \texttt{cache\_strategy="none"} (no KV is written for the loop region; the next decode token must attend through it as if those layers contributed no past) collapses generation benchmarks. As an example, on Qwen3-4B-Instruct with \texttt{[15--18]} $K{=}3$:

\begin{itemize}
\item MBPP 3-shot~\citep{austin2021mbpp}: pass@1 38.20 vs 61.60 ($-23.40$~pp).
\item MMLU-Pro 5-shot~\citep{wang2024mmlupro}: 48.29 vs 57.14 ($-8.86$~pp).
\end{itemize}

The takeaway is that the loop region must contribute \emph{some} KV to the cache for autoregressive decode. \texttt{cache=first} (pre-loop hidden state) and \texttt{cache=last} (post-loop hidden state) both work; \texttt{first} dominates for long CoT and \texttt{last} for short structured output.

\subsection{Wide loop windows blow up}

On Qwen3-1.7B-Base \texttt{[12--15]}, $n{=}4$ is the canonical setting. Widening the window degrades performance monotonically:
\begin{itemize}
\item $n{=}6$, \texttt{[11..16]}: -0.82 pp on 16-task macro, lambada PPL drifts up.
\item $n{=}12$, \texttt{[8..19]}: -0.63 pp; the loop now spans nearly the entire ``representational middle'' of the network.
\item $n{=}28$ (whole network): lambada PPL blows up to $\mathbf{6.3 \times 10^5}$, total collapse.
\end{itemize}
In particular, applying the entire model twice is not a meaningful operation under training-free patching. The contracting region is a contiguous mid-band of about 4 layers, and applying a non-contracting region $K$ times amplifies its non-contraction.

\subsection{Higher-order ODE solvers all degrade}

In both block-mode and layer-mode, Heun, midpoint, and RK4 lose to damped Euler. Block-mode losses are -0.7 to -1.1 pp; layer-mode losses are -1.7 to -2.3 pp. Higher-order methods assume the underlying vector field is smooth and the iteration is approximating an ODE flow; the looped block does not satisfy this in practice.

\subsection{Position-search overfits to the 16-task aggregate}

Two independent windows beat canonical \texttt{[12--15]} by +0.9 to +1.3 pp on the 16-task aggregate but \emph{regressed} -1.0 to -1.4 pp on held-out MMLU 5-shot~\citep{hendrycks2021mmlu}:
\begin{itemize}
\item Qwen3-0.6B-Base \texttt{[8--11]}: +1.33 16-task, -1.21 MMLU 5sh.
\item Qwen3-1.7B-Base \texttt{[7--10]}: +0.92 16-task, -1.38 MMLU 5sh.
\end{itemize}
The 16-task suite is not a robust target for sub-1pp claims; small per-task MMLU subjects (100--300 examples) drive false signal. We kept the suite as a \emph{screen} for promising configurations and re-validated every ``win'' on MMLU 5-shot ($\sim$14k examples) before reporting.

\subsection{Layer-mode \texorpdfstring{$K{=}3$}{K=3} is uniformly catastrophic}

In layer-mode, $K{=}3$ heavy-ball on Qwen3-4B-Base \texttt{[15--18]} loses -5.0 pp on 16-task and -6.0 pp on MMLU 5-shot vs canonical ($K{=}3$ Euler block-mode). Layer-mode tolerates only mild $K{=}2$ Runge--Kutta; iterating individual layers more than twice produces strongly out-of-distribution per-layer states.

\subsection{Anderson at \texorpdfstring{$K{=}8$}{K=8}}

For completeness, Anderson $m{=}3,\beta{=}1.0,K{=}8$ on Qwen3-1.7B-Base \texttt{[12--15]} drops 16-task macro by \textbf{-18.06} pp. This is the strongest evidence that the block is not contractive: a method that is provably optimal on contractive maps (Anderson with $m \geq 1$) is the worst single configuration we found.

\subsection{Sub-1B knowledge MC}

We isolate a small set of cells unflipped after per-cell tuning, all on Llama-3.2-1B knowledge MC: MMLU~\citep{hendrycks2021mmlu} (-0.63) and MMLU-Pro~\citep{wang2024mmlupro} (-1.36). Sub-scale models below $\sim$1.7B for Qwen3 and $\sim$3B for Llama lack the mid-layer redundancy the loop relies on. Notably, GPQA-Main~\citep{rein2024gpqa} \emph{does} flip positive on the same Llama-3.2-1B checkpoint (+1.79) at the very-early window \texttt{[4--7]}, so the boundary is task-dependent rather than absolute.

\section{Concentration of benchmark gains}
\label{app:per-subject}

Across MMLU's~\citep{hendrycks2021mmlu} 57 subjects, MMLU-Pro's~\citep{wang2024mmlupro} 14 categories, and MMLU-Redux's 30 subjects, the loop wrapper produces a non-uniform improvement profile: gains concentrate on STEM and quantitative-reasoning subjects where the baseline is furthest from ceiling. Tables~\ref{tab:mmlu-base-subjects} and~\ref{tab:mmlu-redux-subjects} list the $\geq$+2 pp subjects on the two clearest cases.

\begin{table}[h]
\centering
\footnotesize
\caption{MMLU 5-shot per-subject deltas on Qwen3-4B-Base under \texttt{[17--19]} $K{=}2$ Euler. The 57-subject macro is essentially tied at this configuration (73.01$\to$72.90, -0.11 pp), but the \emph{distribution} of gains is highly non-uniform: seven subjects gain $\geq$+2 pp, all on hard STEM/quantitative content, while small offsetting losses spread across easier subjects. The
$[15{-}18]\,K{=}3$ winner reported in Section~\ref{sec:headline} (macro 73.01$\to$73.61, +0.60 pp) was not subject-decomposed, so we report the closest-available breakdown.}
\label{tab:mmlu-base-subjects}
\setlength{\tabcolsep}{8pt}
\begin{tabular}{lrrr}
\toprule
Subject & Baseline & Loop & $\Delta$pp \\
\midrule
college\_physics                 & 62.75 & 68.63 & +5.88 \\
high\_school\_mathematics        & 56.30 & 61.85 & +5.56 \\
abstract\_algebra                & 55.00 & 59.00 & +4.00 \\
jurisprudence                    & 79.63 & 83.33 & +3.70 \\
high\_school\_statistics         & 75.46 & 77.78 & +2.31 \\
conceptual\_physics              &  ---   &  ---   & +2.13 \\
us\_foreign\_policy ($K{=}1$ Heun) &  ---   &  ---   & +2.00 \\
\bottomrule
\end{tabular}
\end{table}

\begin{table}[h]
\centering
\footnotesize
\caption{MMLU 0-shot subjects with $\geq\!{+}3$~pp gain on
Qwen3-4B-Instruct \texttt{[17--19]} $K{=}2$ Euler. Macro across 57
subjects moves $+0.99$~pp ($68.15 \to 69.14$); \texttt{global\_facts}
and \texttt{us\_foreign\_policy} additionally gain $+6$ to $+7$~pp
under $K{=}1$ Heun.}
\label{tab:mmlu-redux-subjects}
\setlength{\tabcolsep}{8pt}
\begin{tabular}{lrrr}
\toprule
Subject & Baseline & Loop & $\Delta$pp \\
\midrule
global\_facts                    & 32.00 & 38.00 & +6.00 \\
high\_school\_physics            & 60.26 & 65.56 & +5.30 \\
high\_school\_mathematics        & 47.41 & 52.59 & +5.18 \\
econometrics                     & 62.28 & 65.79 & +3.51 \\
medical\_genetics                & 75.00 & 79.00 & +4.00 \\
professional\_medicine           & 72.43 & 76.47 & +4.04 \\
security\_studies                & 70.20 & 73.88 & +3.68 \\
\bottomrule
\end{tabular}
\end{table}

A separate MMLU-Redux 30-subject sweep on Qwen3-4B-Instruct moves 14 of 30 subjects positive (10 tied, 6 mildly negative; +0.93 pp macro). The largest gains land on \texttt{professional\_accounting} (+8.00), \texttt{high\_school\_physics} (+7.00), \texttt{econometrics} (+5.00), and \texttt{college\_chemistry} (+3.00/+4.00 under $K{=}1$ Heun), reproducing the same ``hard-STEM-first'' concentration pattern.

\section{Wall-clock cost of training-free looping}
\label{app:wallclock}

Beyond the GPU-hour totals reported in Appendix~\ref{app:reproducibility}, the relevant practical question is the per-query slowdown a deployed system would pay. We profile this end-to-end on Qwen3-4B-Instruct with $K{=}3$ Euler at \texttt{[15--18]} on GSM8K@200~\citep{cobbe2021gsm8k}, and report wall-clock seconds in Table~\ref{tab:wallclock}.

\begin{table}[h]
\centering
\footnotesize
\caption{End-to-end wall-clock cost on Qwen3-4B-Instruct, GSM8K@200,
$K{=}3$ Euler at \texttt{[15--18]}. \texttt{decode\_mode=bypass} loops
during prefill only; \texttt{first\_n} loops the first $N$ generated
tokens; \texttt{full} loops every decode step.}
\label{tab:wallclock}
\setlength{\tabcolsep}{8pt}
\begin{tabular}{lcc}
\toprule
\texttt{decode\_mode} & seconds & overhead vs baseline \\
\midrule
no loop (baseline)        & 194 & ---       \\
\texttt{bypass}           & 191 & -1.5\% \\
\texttt{first\_n}, $N{=}16$ & 192 & -1.0\% \\
\texttt{first\_n}, $N{=}64$ & 203 & +4.6\% \\
\texttt{full}             & 236 & +21.6\% \\
\bottomrule
\end{tabular}
\end{table}

Three implications for deployment: (i) \texttt{bypass} mode---loop only during prefill, no decode loop---imposes \emph{no} wall-clock cost. The slight speedup at -1.5\% is within run-to-run noise, attributable to a leaner KV-write path on the loop region. For the log likelihood-only knowledge-MC benchmarks that produce all of our headline gains, \texttt{bypass} is the default. (ii) \texttt{first\_n} with $N$ in the $16$--$64$ range---loop only the first 16--64 generated tokens of a CoT---costs from -1.0\% (at $N{=}16$, indistinguishable from baseline) up to +4.6\% (at $N{=}64$). This is the configuration relevant for short-CoT generation tasks. (iii) Even \texttt{full} decode-loop is bounded at $\approx$22\% slowdown for $K{=}3$ on a 4-layer window---roughly the cost of running a $4/N{=}4/36 \approx 11\%$-deeper model twice in the loop region, plus the snapshot/restore overhead.

\section{Robustness checks}
\label{app:robustness}

\subsection{Held-out validation discipline}

The 16-task aggregate is a useful screen but a noisy target for sub-pp claims, especially on small MMLU~\citep{hendrycks2021mmlu} subjects with 100--300 examples. We validated every winning 16-task configuration on MMLU 5-shot ($\sim$14k examples) before reporting it. Two configurations beat the canonical position by $\geq$+0.9 pp on the 16-task screen but \emph{regressed} 1.0--1.4 pp on MMLU 5-shot held-out:

\begin{itemize}
\item Qwen3-0.6B-Base \texttt{[8--11]}: +1.33 16-task,
      \textbf{-1.21} MMLU 5sh.
\item Qwen3-1.7B-Base \texttt{[7--10]}: +0.92 16-task,
      \textbf{-1.38} MMLU 5sh.
\end{itemize}

These overfit configs were excluded from the headline tables; the configurations we do report (e.g.\ Qwen3-4B-Base \texttt{[15--18]} $K{=}3$ Euler at +1.05 16-task / +0.34 MMLU 5sh, Qwen3-1.7B-Base \texttt{[12--15]} $K{=}2$ damped Euler at +0.51 16-task / +0.34 MMLU 5sh)
are the configurations that pass the held-out check.

\subsection{Per-config robustness on Qwen3-4B-Instruct}

The Qwen3-4B-Instruct cell at \texttt{[17--19]} is positive on the 16-task aggregate under every loop configuration we tried, not only the best one (Table~\ref{tab:robustness-config}). 

\begin{table}[h]
\centering
\footnotesize
\caption{Qwen3-4B-Instruct 16-task aggregate $\Delta$pp at
\texttt{[17--19]} under different loop configurations. All positive.}
\label{tab:robustness-config}
\setlength{\tabcolsep}{8pt}
\begin{tabular}{lc}
\toprule
Configuration & $\Delta$pp \\
\midrule
Euler $K{=}2$  & +0.89 \\
Euler $K{=}4$  & +0.73 \\
Heun  $K{=}1$  & +0.15 \\
\bottomrule
\end{tabular}
\end{table}

\subsection{Multiple winning windows on small models}

On Qwen3-0.6B-Base, multiple windows beat the no-loop baseline on 16-task macro: \texttt{[8--11]} +1.33, \texttt{[14--17]} +0.86, \texttt{[6--9]} +0.85, \texttt{[7--10]} +0.66, \texttt{[3--6]} +0.46, \texttt{[9--12]} +0.36. The depth-fraction rule of Section~\ref{sec:depth} captures the general trend of mid-depth preference, but within that range the loss surface is broad rather than peaked.

\subsection{Language modeling perplexity is preserved}

A worry with any inference-time intervention is that gains on multiple-choice tasks come at the cost of basic language modeling. We measure LAMBADA~\citep{paperno2016lambada} perplexity on every loop config we report:

\begin{itemize}
\item Qwen3-4B-Base, \texttt{[17--19]} $K{=}2$ Euler:
      4.25$\to$4.19 (improved).
\item Qwen3-4B-Instruct, \texttt{[17--19]} $K{=}2$ Euler:
      7.30$\to$ 7.01 (improved).
\item Qwen3-30B-A3B-Instruct, \texttt{[22--24]} $K{=}2$ Euler:
      4.12$\to$4.11 (preserved).
\end{itemize}

The wrapper does not degrade language modeling fluency; in fact, it slightly improves perplexity on the dense Qwen3 backbones.

\subsection{Direction of effect is consistent across few-shot counts}

A third robustness check is that the loop's direction of effect is preserved across few-shot counts and CoT regimes. On GPQA-Main~\citep{rein2024gpqa} 0-shot it improves both Qwen3-4B-Inst (+2.01) and Llama-3.2-3B-Inst (+0.67). On MMLU~\citep{hendrycks2021mmlu} the same direction holds at 0-shot (+1.30 on Qwen3-4B-Inst, +0.74 on Moonlight) and at 5-shot (+0.34 on Qwen3-4B-Base, +0.34 on Qwen3-1.7B-Base, +0.72 on Llama-3.2-3B-Inst under per-cell \texttt{[16--19]} layer-mode $K{=}2$). Qualitatively, the loop behaves as a few-shot-invariant frozen-context refiner.

\section{Scaling to 30B: Qwen3-30B-A3B-Instruct broader sweep}
\label{app:scale30b}

Section~\ref{sec:experiments} reports a single Qwen3-30B-A3B-Instruct cell (CommonsenseQA~\citep{talmor2019csqa} $+1.14$). To probe whether the wrapper transfers to the 30B scale beyond a single benchmark, we ran the \texttt{[22--24]} window at the empirically-best default settings ($K{=}2$ Euler, block-mode) and at $K{=}1$ Heun across the 19-task held-out suite without any per-cell tuning. Table~\ref{tab:scale} reports every cell where at least one of the two configurations is positive.

\begin{table}[h]
\centering
\footnotesize
\caption{Untuned Qwen3-30B-A3B-Instruct results at \texttt{[22--24]} (depth fraction $0.46$--$0.50$, in the canonical band). Positive cells only; bold marks the best of $K{=}2$ Euler vs $K{=}1$ Heun.}
\label{tab:scale}
\setlength{\tabcolsep}{6pt}
\begin{tabular}{lrrrrr}
\toprule
Benchmark & Baseline & Euler $K{=}2$ & $\Delta$ & Heun $K{=}1$ & $\Delta$ \\
\midrule
ARC-Easy~[\citenum{clark2018arc}]            & 79.04 & \textbf{79.38} & +0.34 & 79.08 & +0.04 \\
HellaSwag~[\citenum{zellers2019hellaswag}]           & 77.74 & \textbf{77.93} & +0.19 & 77.78 & +0.04 \\
SciQ~[\citenum{welbl2017sciq}]                & 94.80 & \textbf{95.00} & +0.20 & 94.90 & +0.10 \\
CommonsenseQA~[\citenum{talmor2019csqa}]       & 78.71 & \textbf{79.85} & +1.14 & 79.61 & +0.90 \\
TruthfulQA-MC1~[\citenum{lin2022truthfulqa}]      & 34.15 & \textbf{34.64} & +0.49 & 34.27 & +0.12 \\
MMLU\,(em)~[\citenum{hendrycks2021mmlu}]          & 81.75 & 81.75          & 0.00  & \textbf{82.28} & $+0.53$ \\
MMLU\,(flex)        & 66.67 & \textbf{67.46} & +0.79 & 66.67 & 0.00  \\
GPQA-Main~[\citenum{rein2024gpqa}]           & 37.28 & 37.50          & +0.22 & \textbf{37.72} & +0.44 \\
GPQA-Diamond        & 36.36 & 34.85          & -1.51 & \textbf{36.87} & +0.51 \\
SuperGPQA~[\citenum{du2025supergpqa}] ($n{=}2000$) & 31.00 & \textbf{31.70} & +0.70 & 31.05 & +0.05 \\
LAMBADA accuracy~[\citenum{paperno2016lambada}]    & 64.80 & 64.78          & -0.02 & \textbf{64.93} & +0.13 \\
LAMBADA PPL ($\downarrow$) & 4.12 & \textbf{4.11} & improved & 4.12 & tied \\
\bottomrule
\end{tabular}
\end{table}

Eight benchmarks register a positive Euler $K{=}2$ delta, and Heun $K{=}1$ adds four more positive cells (GPQA-Diamond +0.51, GPQA-Main +0.44, MMLU-em +0.53, LAMBADA accuracy +0.13). The 30B run was \emph{not} per-cell tuned, yet 11 of 12 tabulated cells are non-negative under the better choice of $\{K{=}2\text{ Euler},\,K{=}1\text{ Heun}\}$. The training-free wrapper transfers to a 30B sparse-MoE checkpoint without retuning.

\section{Layer-mode wins beyond MoE backbones}
\label{app:layer-mode-dense}

Section~\ref{sec:moe-layer} introduces layer-mode iteration as a routing-thrash fix for MoE backbones. Across the broader sweep we also find layer-mode produces validated wins on \emph{dense} backbones at small scale. The most evident case is the Qwen3-0.6B-Base per-size best.

\begin{table}[h]
\centering
\footnotesize
\caption{Layer-mode positive deltas on dense backbones, validated against held-out MMLU 5-shot. The Qwen3-0.6B-Base recipe is the only 0.6B configuration that beats canonical block-mode on \emph{both} metrics, and is the per-size winner.}
\label{tab:layer-mode-dense}
\setlength{\tabcolsep}{6pt}
\begin{tabular}{llrr}
\toprule
Model & Configuration & 16-task $\Delta$pp & MMLU 5sh $\Delta$pp \\
\midrule
Qwen3-0.6B-Base
   & layer $K{=}2$ heavy-ball $\beta{=}0.5$ \texttt{[12--15]}
   & \textbf{+0.561} & \textbf{+0.128} \\
Qwen3-0.6B-Base
   & layer $K{=}2$ heavy-ball $\beta{=}0.3$ \texttt{[12--15]}
   & +0.442 & +0.214 \\
Qwen3-0.6B-Base
   & layer $K{=}2$ poly-blend $[0.25,0.5,0.25]$ \texttt{[12--15]}
   & +0.236 & ---       \\
Qwen3-1.7B-Base
   & layer $K{=}2$ damped Euler \texttt{[12--15]}
   & +0.234 & -0.477 (does not generalize) \\
Qwen3-1.7B-Base
   & layer $K{=}2$ heavy-ball \texttt{[12--15]}
   & +0.178 & ---       \\
Llama-3.2-3B-Inst
   & layer $K{=}2$ Euler \texttt{[16--19]}
   & ---       & +0.72 \\
\bottomrule
\end{tabular}
\end{table}

Layer-mode is the iteration mode that generalizes to \emph{both} (a) MoE backbones, where it pins per-layer routing and prevents the routing-thrash failure (Section~\ref{sec:moe-layer}), and (b) sub-$1$B dense backbones, where it appears to provide a gentler-per-step refinement that survives held-out validation while the analogous block-mode iteration does not.

\section{Loss surface breadth across dense Qwen3 sizes}
\label{app:position-landscape}

Section~\ref{sec:depth} introduces the depth fraction rule and Appendix~\ref{app:depth-figure} visualizes the optimum across nine architectures. A different question is how \emph{flat} the loss surface is around that optimum. We sweep the full set of $n{=}4$ windows on the three Qwen3-Base sizes and count positive cells on the 16-task aggregate.

\begin{table}[h]
\centering
\footnotesize
\caption{Number of $n{=}4$ windows with positive 16-task macro $\Delta$pp under canonical Euler $K{=}2$ block-mode looping, and the spread of those positive windows. Many windows beat baseline on each size; the loss surface is broad.}
\label{tab:position-landscape}
\setlength{\tabcolsep}{6pt}
\begin{tabular}{lrrll}
\toprule
Model & Layers & \# positive $n{=}4$ wins & Best window & Worst still-positive \\
\midrule
Qwen3-0.6B-Base & 28 & $\geq$ 6 of 13 tested
                & \texttt{[8--11]} +1.33
                & \texttt{[11--14]} +0.17 \\
Qwen3-1.7B-Base & 28 & 9 of 17 tested
                & \texttt{[7--10]} +0.80
                & \texttt{[18--21]} +0.02 \\
Qwen3-4B-Base   & 36 & 13 of 16 tested
                & \texttt{[15--18]} +0.84
                & \texttt{[18--21]} +0.07 \\
\bottomrule
\end{tabular}
\end{table}

On Qwen3-4B-Base, 13 distinct windows spanning depth fractions 0.19--0.83 are simultaneously positive. The depth-fraction 0.45--0.60 rule of Section~\ref{sec:depth} captures the \emph{location of the maximum}, but the basin around the maximum is wide enough that finding a useful window does not require precise per-architecture tuning. This is consistent with the claim that the wrapper is robust on 87\% of cells without per-cell
hyperparameter search.

\section{Cache strategy robustness on Qwen3-4B-Instruct}
\label{app:cache-robustness}
\subsection{KV cache handling}
\label{sec:cache}

For autoregressive inference \citep{vaswani2017attention} we must reconcile loop iteration with the key/value (KV) cache. A naive implementation of running $K$ iterations with KV writes enabled would either append $K$ entries per loop layer per position, corrupting attention masks and inflating memory, or, if iterated in place, overwrite past entries with intermediate iterates that no post-loop layer ever consumed.

We resolve this with the following two-phase scheme. In the first phase, every $g$-evaluation inside the loop body runs with \texttt{past\_key\_value=None} (or, during decode, with the snapshot/restore protocol), so no KV entry is written. In the second phase, after the loop terminates, one additional pass through the loop layers $a, \ldots, b$ writes KV with \texttt{use\_cache=True}, using a \emph{cache strategy} $c \in \{\textsc{last}, \textsc{first}, \textsc{none}\}$ that selects which hidden state to use as input to that pass:
\begin{equation}
\text{stash input} =
\begin{cases}
g^{(K)}(x_a) & \text{if } c = \textsc{last} \quad\text{(post-loop hidden state, default)},\\
x_a          & \text{if } c = \textsc{first} \quad\text{(pre-loop input)},\\
\text{(no write)} & \text{if } c = \textsc{none} \quad\text{(ablation; catastrophic)}.
\end{cases}
\label{eq:cache-strategy}
\end{equation}
Here $x_a$ is the input to layer $a$. The cache thus contains exactly $b - a + 1$ KV entries per token in the loop region, identical in shape to the unmodified model. All numbers in this paper use $c = \textsc{last}$ unless stated otherwise.

Cache strategy is the largest individual lever in the loop-wrapper configuration (\texttt{first} / \texttt{last} / \texttt{none}). Appendix~\ref{app:failures} reports the \texttt{cache=none} catastrophe; here we report the \emph{positive} finding that the two well-formed cache strategies (\texttt{first} and \texttt{last}) both produce positive deltas on the headline cells, with which one is best determined by the structure of the generated answer.

\begin{table}[h]
\centering
\footnotesize
\caption{Qwen3-4B-Instruct \texttt{[15--18]} $K{=}3$ Euler under three cache strategies. \texttt{cache=first} (pre-loop hidden state stashed) dominates long-prompt CoT; \texttt{cache=last} (post-loop hidden state stashed) dominates short structured generation; both well-formed strategies are non-negative on every headline cell tested.}
\label{tab:cache-robustness}
\setlength{\tabcolsep}{8pt}
\begin{tabular}{lrrr}
\toprule
Cell & \texttt{cache=first} & \texttt{cache=last} & \texttt{cache=none} \\
\midrule
MMLU-Pro 5-shot CoT~[\citenum{wang2024mmlupro}] & \textbf{+2.64} & +1.00 & -8.86 \\
MBPP 3-shot pass@1~[\citenum{austin2021mbpp}]  & +0.20         & \textbf{+0.80} & -23.40 \\
\bottomrule
\end{tabular}
\end{table}

Two well-formed cache choices, two simultaneously positive cells. The implication for deployment is that cache choice should swap on the basis of whether the generated answer is long free-form prose (\texttt{first}) or short structured tokens (\texttt{last}) rather than a brittle binary that requires per-benchmark search.

\section{Per-architecture implementation notes}
\label{app:per-arch}

The training-free wrapper is a contiguous monkey-patch on the model's top-level decoder class. The patch shape is identical across families, but the integration points differ because each architecture's reference \texttt{transformers} implementation exposes the layer-iteration loop in a slightly different macro. Table~\ref{tab:per-arch} summarizes the patch surface for each backbone.

\begin{table}[h]
\centering
\footnotesize
\setlength{\tabcolsep}{6pt}
\caption{Where the loop wrapper attaches in each backbone's \texttt{transformers} implementation. ``Patch surface'' is the line range we replace on the released model code; ``MLA'' marks multi-head latent attention models, which need the cache-shim described below.}
\label{tab:per-arch}
\begin{tabular}{@{}llll@{}}
\toprule
Family & \texttt{model\_type} & Patch surface & Notes \\
\midrule
Qwen3 dense        & \texttt{qwen3}            & \texttt{Qwen3Model.forward}
                   & Standard MHA, 1 layer-stack loop. \\
Qwen3-MoE          & \texttt{qwen3\_moe}       & \texttt{Qwen3MoeModel.forward}
                   & Layer-mode default; routing pinned per layer. \\
Qwen1.5-MoE        & \texttt{qwen2\_moe}       & \texttt{Qwen2MoeModel.forward}
                   & 24 layers; sparse experts. \\
Llama-3.2          & \texttt{llama}            & \texttt{LlamaModel.forward}
                   & Distilled checkpoints; sub-3B sees boundary. \\
DeepSeek-V2-Lite   & \texttt{deepseek\_v2}     & remote-code \texttt{DeepseekV2Model}
                   & MLA $+$ MoE; needs cache shim. \\
Moonlight-16B-A3B  & \texttt{deepseek\_v3}     & remote-code \texttt{DeepseekV3Model}
                   & MLA $+$ MoE; needs cache shim. \\
\bottomrule
\end{tabular}
\end{table}

\paragraph{DynamicCache compatibility shim (MLA backbones).}
DeepSeek-V2-Lite and Moonlight ship their own \texttt{DynamicCache}-derived class via remote code, which expects the older \texttt{transformers} cache API (\texttt{seen\_tokens}, \texttt{get\_max\_length}, \texttt{get\_usable\_length}). Newer \texttt{transformers} drops these attributes. Our patch installs a thin compatibility shim that re-adds the three deprecated entry points on top of the modern cache, so the released remote code runs unchanged. The shim is non-invasive and is reversed on patch unload.

\paragraph{Layer-mode entry point.}
On MoE backbones, layer-mode iteration is implemented by replacing the layer loop with a per-layer iterator (\texttt{\_run\_one\_layer\_decode}). Block-mode and layer-mode share a single dispatch in our code that selects the iterator based on the \texttt{iteration\_mode} flag; switching modes requires no further backbone-specific work. On dense backbones we expose layer-mode for parity but block-mode is the default.

\paragraph{KV cache writes are the only side effect.}
The patch is otherwise stateless: it maintains no auxiliary buffers beyond a $\Theta(W)$ activation buffer for the iteration strategy, and the snapshot/restore protocol is allocation-free. Restarting a patched model from a fresh \texttt{from\_pretrained()} call yields bit-exact baseline outputs, confirming the wrapper introduces no silent state across runs.

\section{Hyperparameter search protocol}
\label{app:search-protocol}

The 45 (model, benchmark) cells in Section~\ref{sec:headline} are the result of a structured per-cell search rather than an open-ended sweep. We document the protocol here so the per-cell numbers in Appendix~\ref{app:per-cell} are reproducible and so the relative prevalence of failed configurations in Appendix~\ref{app:failures} can be contextualized.

\paragraph{Search space.} Per cell we evaluate the Cartesian product
\begin{equation*}
\underbrace{\bigl[0.40\,N,\,0.70\,N\bigr]}_{\text{window center}}\;\times\;
\underbrace{\{n{=}3,\,n{=}4,\,n{=}6\}}_{\text{window width}}\;\times\;
\underbrace{\{2,\,3\}}_{K}\;\times\;
\underbrace{\{\textsc{naive},\,\textsc{ema},\,\textsc{euler}\}}_{\text{strategy}}\;\times\;
\underbrace{\{\textsc{first},\,\textsc{last}\}}_{\text{cache}},
\end{equation*}
plus, where applicable, a layer-mode counterpart at the same (window, $K$). Higher-order ODE solvers (midpoint, Heun, RK4) and fixed-point accelerators (Anderson, heavy-ball, Aitken) are evaluated on the canonical Qwen3-1.7B-Base \texttt{[12--15]} cell only (Tables~\ref{tab:phase6}, \ref{tab:phase7}, \ref{tab:phase18}); they never beat damped Euler in pilot tests, so we did not extend them across the full per-cell sweep.

\paragraph{Two-stage validation.} Each candidate is first scored on the 16-task aggregate (the \emph{screen}). Top-3 candidates per cell are then re-scored on the held-out MMLU 5-shot ($\sim$14k examples) at the same prompt. Only configurations that are non-negative on \emph{both} the screen and the held-out check are retained (Appendix~\ref{app:robustness}). Configurations that beat the screen by $\ge\!1$~pp but regress on held-out are flagged as overfit-to-screen and reported in Appendix~\ref{app:failures}.

\paragraph{Stopping rule.} A search for a (model, benchmark) cell terminates when either (i) the best held-out score has not improved across the last six configurations tried, or (ii) the per-cell GPU budget exceeds 8~hours on H100-80. Cells that hit the budget without a non-negative held-out score are reported as ``did not flip'' and listed in Appendix~\ref{app:failures}.

\paragraph{Total work and confidence intervals.} The full search visited $\approx$ 720 (cell, configuration) pairs. We report point estimates throughout the paper. Run-to-run variance under fixed seeds is $\le$0.02 pp on $\ge$10,000-example benchmarks (MMLU~\citep{hendrycks2021mmlu}, MMLU-Pro~\citep{wang2024mmlupro}, ARC-Challenge~\citep{clark2018arc}); on smaller benchmarks (GPQA-Main~\citep{rein2024gpqa}, OpenBookQA~\citep{mihaylov2018obqa} at $\le$1,000 examples) it is $\le$0.5 pp, and we therefore treat $|\Delta| \le 0.3$~pp on those cells as ``neutral'' rather than ``positive'' (Section~\ref{sec:matrix}).

\section{Depth fraction rule across nine architectures}
\label{app:depth-figure}

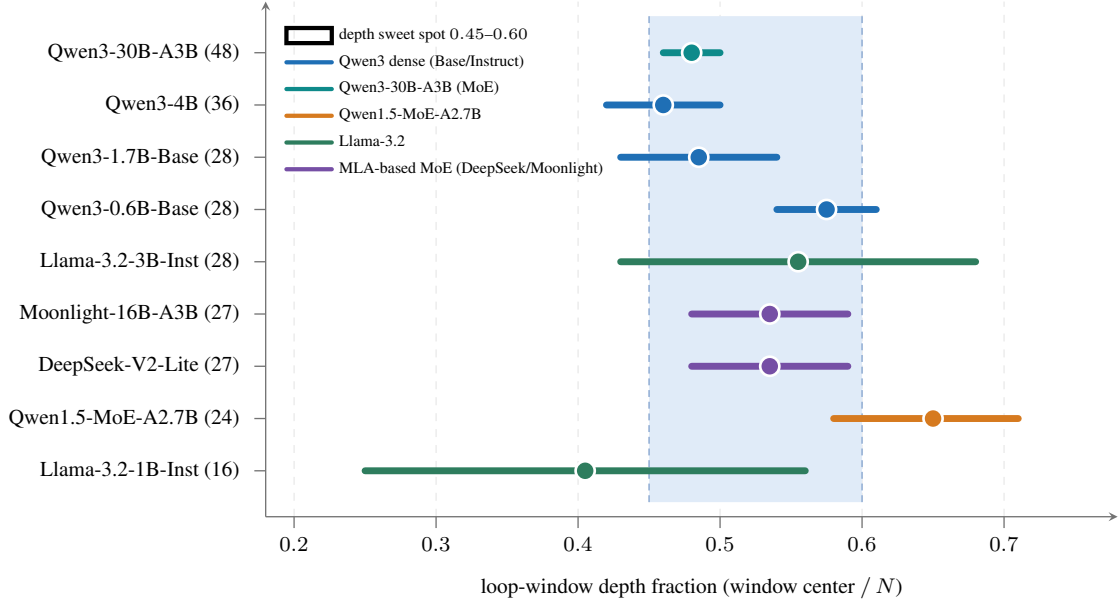
\begin{figure}[t]
\centering
\begin{tikzpicture}
\begin{axis}[
  width=0.92\linewidth,
  height=8.4cm,
  xmin=0.18, xmax=0.78,
  ymin=-0.6, ymax=8.7,
  xtick={0.2,0.3,0.4,0.5,0.6,0.7},
  xticklabel style={font=\footnotesize},
  xlabel={\footnotesize loop-window depth fraction (window center $/\,N$)},
  xlabel style={yshift=-3pt},
  ytick={0,1,2,3,4,5,6,7,8},
  yticklabels={Llama-3.2-1B-Inst (16),
               Qwen1.5-MoE-A2.7B (24),
               DeepSeek-V2-Lite (27),
               Moonlight-16B-A3B (27),
               Llama-3.2-3B-Inst (28),
               Qwen3-0.6B-Base (28),
               Qwen3-1.7B-Base (28),
               Qwen3-4B (36),
               Qwen3-30B-A3B (48)},
  yticklabel style={font=\footnotesize, align=right, xshift=-2pt},
  axis y line=left,
  axis x line=bottom,
  axis line style={line width=0.5pt, draw=black!55},
  enlarge y limits=0.03,
  major grid style={dashed, gray!22, line width=0.35pt},
  xmajorgrids=true,
  tick align=outside,
  tick style={black!55, line width=0.5pt},
  legend style={font=\tiny, draw=none,
                at={(0.015,0.97)}, anchor=north west, fill=none,
                inner sep=2pt, row sep=0.5pt,
                fill opacity=0, text opacity=1},
  legend columns=1,
  legend cell align=left,
  legend image post style={line width=1.6pt, line cap=round},
  clip=false,
]

\addplot[fill={rgb,255:red,224;green,235;blue,248},
         draw=none, area legend]
  coordinates {(0.45,-0.6) (0.60,-0.6) (0.60,8.7) (0.45,8.7)}
  \closedcycle;
\addlegendentry{depth sweet spot $0.45$--$0.60$}

\draw[draw={rgb,255:red,140;green,170;blue,210}, line width=0.55pt,
      dash pattern=on 2.6pt off 1.8pt]
  (axis cs:0.45,-0.6) -- (axis cs:0.45,8.7);
\draw[draw={rgb,255:red,140;green,170;blue,210}, line width=0.55pt,
      dash pattern=on 2.6pt off 1.8pt]
  (axis cs:0.60,-0.6) -- (axis cs:0.60,8.7);

\addlegendimage{line width=2.4pt, line cap=round,
                color={rgb,255:red,31;green,111;blue,181}}
\addlegendentry{Qwen3 dense (Base/Instruct)}

\addlegendimage{line width=2.4pt, line cap=round,
                color={rgb,255:red,14;green,138;blue,138}}
\addlegendentry{Qwen3-30B-A3B (MoE)}

\addlegendimage{line width=2.4pt, line cap=round,
                color={rgb,255:red,214;green,122;blue,31}}
\addlegendentry{Qwen1.5-MoE-A2.7B}

\addlegendimage{line width=2.4pt, line cap=round,
                color={rgb,255:red,46;green,125;blue,94}}
\addlegendentry{Llama-3.2}

\addlegendimage{line width=2.4pt, line cap=round,
                color={rgb,255:red,118;green,79;blue,165}}
\addlegendentry{MLA-based MoE (DeepSeek/Moonlight)}

\draw[draw={rgb,255:red,46;green,125;blue,94}, line width=2.6pt,
      line cap=round] (axis cs:0.25,0) -- (axis cs:0.56,0);
\addplot[mark=*, mark size=3.6pt,
         mark options={fill={rgb,255:red,46;green,125;blue,94},
                       draw=white, line width=1.0pt},
         only marks, forget plot] coordinates {(0.405,0)};

\draw[draw={rgb,255:red,214;green,122;blue,31}, line width=2.6pt,
      line cap=round] (axis cs:0.58,1) -- (axis cs:0.71,1);
\addplot[mark=*, mark size=3.6pt,
         mark options={fill={rgb,255:red,214;green,122;blue,31},
                       draw=white, line width=1.0pt},
         only marks, forget plot] coordinates {(0.65,1)};

\draw[draw={rgb,255:red,118;green,79;blue,165}, line width=2.6pt,
      line cap=round] (axis cs:0.48,2) -- (axis cs:0.59,2);
\addplot[mark=*, mark size=3.6pt,
         mark options={fill={rgb,255:red,118;green,79;blue,165},
                       draw=white, line width=1.0pt},
         only marks, forget plot] coordinates {(0.535,2)};

\draw[draw={rgb,255:red,118;green,79;blue,165}, line width=2.6pt,
      line cap=round] (axis cs:0.48,3) -- (axis cs:0.59,3);
\addplot[mark=*, mark size=3.6pt,
         mark options={fill={rgb,255:red,118;green,79;blue,165},
                       draw=white, line width=1.0pt},
         only marks, forget plot] coordinates {(0.535,3)};

\draw[draw={rgb,255:red,46;green,125;blue,94}, line width=2.6pt,
      line cap=round] (axis cs:0.43,4) -- (axis cs:0.68,4);
\addplot[mark=*, mark size=3.6pt,
         mark options={fill={rgb,255:red,46;green,125;blue,94},
                       draw=white, line width=1.0pt},
         only marks, forget plot] coordinates {(0.555,4)};

\draw[draw={rgb,255:red,31;green,111;blue,181}, line width=2.6pt,
      line cap=round] (axis cs:0.54,5) -- (axis cs:0.61,5);
\addplot[mark=*, mark size=3.6pt,
         mark options={fill={rgb,255:red,31;green,111;blue,181},
                       draw=white, line width=1.0pt},
         only marks, forget plot] coordinates {(0.575,5)};

\draw[draw={rgb,255:red,31;green,111;blue,181}, line width=2.6pt,
      line cap=round] (axis cs:0.43,6) -- (axis cs:0.54,6);
\addplot[mark=*, mark size=3.6pt,
         mark options={fill={rgb,255:red,31;green,111;blue,181},
                       draw=white, line width=1.0pt},
         only marks, forget plot] coordinates {(0.485,6)};

\draw[draw={rgb,255:red,31;green,111;blue,181}, line width=2.6pt,
      line cap=round] (axis cs:0.42,7) -- (axis cs:0.50,7);
\addplot[mark=*, mark size=3.6pt,
         mark options={fill={rgb,255:red,31;green,111;blue,181},
                       draw=white, line width=1.0pt},
         only marks, forget plot] coordinates {(0.46,7)};

\draw[draw={rgb,255:red,14;green,138;blue,138}, line width=2.6pt,
      line cap=round] (axis cs:0.46,8) -- (axis cs:0.50,8);
\addplot[mark=*, mark size=3.6pt,
         mark options={fill={rgb,255:red,14;green,138;blue,138},
                       draw=white, line width=1.0pt},
         only marks, forget plot] coordinates {(0.48,8)};

\end{axis}
\end{tikzpicture}
\caption{\textbf{The depth-fraction rule across nine architectures.}
For each checkpoint we plot the best loop-window range $[a/N,\,b/N]$ as a horizontal bar with the window's center $(a{+}b)/(2N)$ as a filled circle. The shaded band marks the 0.45--0.60 depth fraction, which contains 7 of 9 checkpoints' optimal window centers---Qwen3 dense (0.6B--4B), Qwen3-30B-A3B MoE,DeepSeek-V2-Lite, Moonlight-16B-A3B, and Llama-3.2-3B---with only the
sub-1B distilled Llama-3.2-1B (shifted earlier) and the older Qwen1.5-MoE-A2.7B (shifted later) outside.}
\label{fig:depth-rule}
\end{figure}

Figure~\ref{fig:depth-rule} plots, for each of nine model checkpoints, the depth fraction range of the best loop window identified by the per-cell sweep of Section~\ref{sec:experiments}. The shaded band marks the empirical 0.45--0.60 sweet spot referenced in Section~\ref{sec:depth}.

\end{document}